%% file: preprint.tex
\documentclass{article} % For LaTeX2e
\usepackage{colm2024_conference}

\input{config}
\input{math_commands}

\newcommand{\model}{\textsc{ExoViP}\xspace}
\newcommand{\visprog}{\textsc{VisProg}\xspace}

\newcommand{\code}[1]{{\fontfamily{lmss}\selectfont #1}}

\title{ExoViP: Step-by-step Verification and Exploration with \\ Exoskeleton Modules for Compositional Visual Reasoning}

\colmfinalcopy

\makeatletter
\def\thanks#1{\protected@xdef\@thanks{\@thanks
        \protect\footnotetext{#1}}}
\makeatother

\author{Yuxuan Wang$^{\,1}$\quad Alan Yuille$^{\,2}$\quad Zhuowan Li$^{\,2\,\textrm{\Letter}}$\thanks{\hspace{-3mm}\textsuperscript{\Letter}~Corresponding authors.}\quad Zilong Zheng$^{\,1\,\textrm{\Letter}}$ \\ 
$^1$ State Key Laboratory of General Artificial Intelligence, BIGAI, Beijing, China \\
$^2$ Johns Hopkins University, Baltimore, Maryland, USA \\
\texttt{wangyuxuan1@bigai.ai}\qquad\texttt{\{ayuille1,zli110\}@jhu.edu}\qquad\texttt{zlzheng@bigai.ai}
}

\begin{document}

% \vspace{-2em}
% \maketitle

\maketitle

\begin{abstract}

Compositional visual reasoning methods, which translate a complex query into a structured composition of feasible visual tasks, have exhibited a strong potential in complicated multi-modal tasks. Empowered by recent advances in large language models~(LLMs), this multi-modal challenge has been brought to a new stage by treating LLMs as few-shot/zero-shot planners, \ie, vision-language~(VL) programming~\citep{Gupta_2023_CVPR}.
Such methods, despite their numerous merits, suffer from challenges due to LLM planning mistakes or inaccuracy of visual execution modules, lagging behind the non-compositional models.
In this work, we devise a ``plug-and-play'' method, \model, to correct errors in both the planning and execution stages through introspective verification. We employ verification modules as ``exoskeletons'' to enhance current VL programming schemes. Specifically, our proposed verification module utilizes a mixture of three sub-verifiers to validate predictions after each reasoning step, subsequently calibrating the visual module predictions and refining the reasoning trace planned by LLMs. 
Experimental results on two representative VL programming methods showcase consistent improvements on five compositional reasoning tasks on standard benchmarks. In light of this, we believe that \model can foster better performance and generalization on open-domain multi-modal challenges.

\begin{center}
    \begin{tabular}{ccl}
        \faGithub & \textbf{Code} & \url{https://github.com/bigai-nlco/ExoViP} \\
        % \faGlobe & \textbf{Webpage} & \url{https://biga-nlco.github.io/ExoViP} 
    \end{tabular}
\end{center}

\end{abstract}

\section{Introduction}

\begin{figure}[th!]
    \centering
    % \vspace{-2.5em}
    \includegraphics[width=\linewidth]{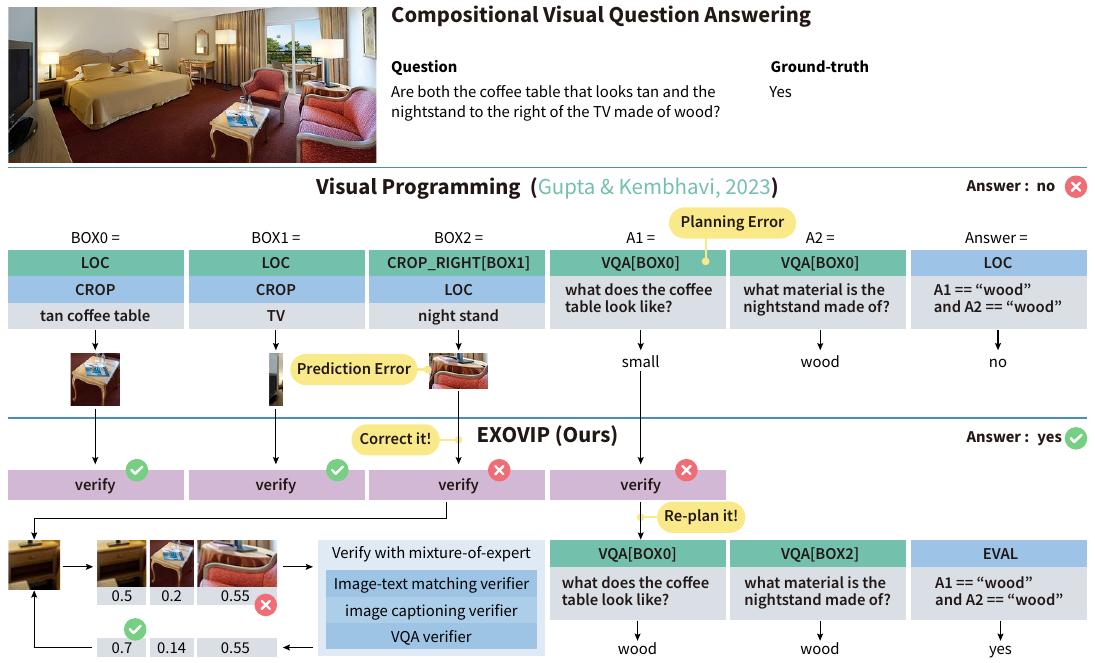}
    % \vspace{-0.5em}
    \caption{\textbf{An overview of \model.} The prediction after each step is verified by the proposed  ``exoskeleton'' verification modules, which contain a mix of three sub-verifiers. The verified scores help correct the errors in the vision module predictions or refine the reasoning programs planned by LLM.}
    \vspace{-1.0em}
    \label{fig:framework}
\end{figure}

%%%%%%%%%%%%%% background

Compositional visual reasoning methods, due to their interpretability and generalization over complex vision-language~(VL) challenges that demand intricate, multi-step visual reasoning guided by linguistic input, have long been the focus for many researchers. 
Traditional compositional techniques, exemplified by neural modular networks~\citep{Andreas2015NeuralMN, DBLP:conf/iccv/HuARDS17, Johnson2017InferringAE, DBLP:conf/eccv/HuADS18,DBLP:conf/naacl/LeCH22,DBLP:conf/mm/Qian0DC022}, have shown success in breaking down intricate language instructions into manageable visual tasks. However, they tend to falter when confronted with the need for broader generalization across diverse domains. Furthermore, the limitations of these approaches manifest in their inability to enhance the interaction and attention between neural modules through supervision or feedback mechanisms, thereby constraining performance to end-to-end training paradigms. Recent advances in large language models (LLM)~\citep{Radford2018ImprovingLU, Radford2019LanguageMA, Brown2020LanguageMA, OpenAI2023GPT4TR, Chowdhery2022PaLMSL} have led to novel methods that harness LLM as zero-shot or few-shot planners to address visual reasoning tasks, notably vision-language programming~(\visprog)~\citep{Gupta_2023_CVPR} and ViperGPT~\citep{surismenon2023vipergpt}. These approaches make use of readily available pretrained vision models and systematically assemble them, guided by the reasoning trace provided by LLMs, resulting in interpretable intermediary outcomes and highly adaptable reasoning capabilities.

%%%%%%%%%%%%%%%% motivation
Despite their merits, current visual programming approaches encounter persistent challenges, often resulting from shortcomings in the planning processes of LLMs or the capabilities of visual modules. More precisely, they often fall short of the performance achieved by non-compositional models; refer to \cref{fig:framework} for exemplar error cases. To investigate these limitations, a manual examination of 100 randomly selected failure cases (\textbf{Section~\ref{supp:failure-case}}) of \visprog~\citep{Gupta_2023_CVPR} on the GQA dataset~\citep{Hudson2019GQAAN} for visual question answering was conducted. The analysis revealed two primary failure categories: Firstly, approximately 30\% of these failures were attributed to \textbf{planning errors}, where the LLM failed to parse the language query into programs correctly, preventing the formulation of a solvable program. Secondly, over 40\% of the failures were attributed to \textbf{module errors}, as the visual modules were incapable of executing the program accurately. The remaining failure cases (less than 30\%) stemmed from issues like synonym usage (\eg, ``woman'' \emph{vs}\onedot ``lady'') or question ambiguity.

Motivated by these failure modes, in this work, we introduce \model, a ``plug-and-play'' method that uses ``exoskeleton'' verification modules to verify the reasoning results step by step, thus correcting the module errors and refining the LLM planning traces. As depicted in~\cref{fig:framework}, \model effectively rectifies both types of errors: the verification module contains a mixture of three sub-verifiers, including an image-text matching verifier, an image captioning verifier, and a visual question answering~(VQA) verifier. These sub-verifiers meticulously validate the accuracy of the predictions generated by the visual modules, thereby correcting module errors. 
For refining the planning traces, a reasoning trace tree is constructed based on the verification scores, along with the self-correctness score~\citep{Pan2023AutomaticallyCL} obtained from LLMs. The methodology involves searching through the tree to identify the optimal trace with the highest score.

% details
% An overview of \model is shown in \cref{fig:framework}. Given a query, we adopt the LLM to plan the solution step by step. At each step, we take exoskeleton modules to calculate the verification scores of the candidate predictions from the program. Inspired by recent advances in multi-modal counterfactual reasoning~\citep{2019mmcf, 2022cfclip}, we construct counterfactual samples of candidate predictions through its antonymies, which are obtained by CLIP~\citep{Radford2021LearningTV}. We further take the counterfactual samples to update the verification scores, which are used to calibrate the candidate predictions. After getting the verification scores, we build a tree-like trace-searching algorithm to improve the planning. Specifically, we choose the next planning step with a higher verification score. Inspired by the strong , we further refine our trace-searching algorithm by LLM itself. To be more specific, we take LLM to select the best next planning step from candidate planning steps sharing similar highest verification scores. 

% contribution
To demonstrate the effectiveness of \model, we apply our method to two recent visual programming methods: self-defined programs, \ie, \visprog~\citep{Gupta_2023_CVPR} and Python code programs, \ie, ViperGPT~\citep{surismenon2023vipergpt}. Our experiments encompass six compositional visual reasoning tasks, including compositional image question answering, referring expression understanding, natural language for visual reasoning, visual abstract reasoning, language-guided image editing, and spatial-temporal reasoning. The experimental results consistently indicate notable improvements in performance for both models. In light of this, we believe \model can foster better performance on open-world compositional reasoning tasks.

In summary, our main contributions are as follows: 
% First, we introduce the ``exoskeleton'' verification modules for visual programming, which verifies the correctness of vision module predictions step by step. Second, we show how the verification modules are leveraged to correct the module errors by calibrating the module predictions, and to correct the planning errors by tree searching considering both verification scores and LLM self-correctness. Third, we apply our method on two models and and show consistent improvements over five tasks, showing the effectiveness of \model.
\begin{itemize}[leftmargin=*, topsep=0pt, noitemsep]
    \item We introduce ``exoskeleton'' verification modules tailored for module error and plan error in existing compositional visual reasoning methods, which systematically validate the accuracy of vision module predictions in a step-by-step manner.
    \item We illustrate the synergistic integration of our proposed verification modules with a tree-based search algorithm, enhanced by the self-correcting capabilities of the LLM. This collaborative design effectively tackles both the module error and plan error. The tree-based search is informed by a verification score, which serves as a measure of confidence in the search process. Concurrently, this verification score is dynamically refined as the search progresses, ensuring a more accurate and reliable verification process.
    % \item We demonstrate how the proposed verification modules are effectively employed to rectify module errors by fine-tuning their predictions and to correct planning errors through a tree search process, taking into account both verification scores and the self-correctness of LLMs.
    \item We have implemented our methodology within two compositional methods, and the outcome has been a uniform enhancement in performance across six diverse tasks encompassing both image and video modalities. This underscores the efficacy of \model in augmenting visual reasoning skills.
    % \item We apply our methodology to two compositonal methods, resulting in consistent performance improvements across five distinct tasks and two modalities, i.e. image and video, thus highlighting the effectiveness of \model in enhancing visual reasoning capabilities.
\end{itemize}

\section{Related work}

\paragraph{LLMs in multi-modal tasks.}
LLMs have significantly enhanced multi-modal tasks through their adaptability and extensive knowledge. There are three primary methods for applying LLMs to multi-modal challenges. One approach involves integrating extra parameters into LLMs for multi-modal contexts and then fine-tuning with either a fixed~\citep{Tsimpoukelli2021multimodalFL,DBLP:conf/nips/AlayracDLMBHLMM22,li2023blip,DBLP:journals/corr/abs-2304-15010,DBLP:journals/corr/abs-2305-03726,Dai2023InstructBLIPTG,DBLP:journals/corr/abs-2306-17107} or an adjustable LLM~\citep{Hao2022LanguageMA,DBLP:journals/corr/abs-2302-14045,DBLP:journals/corr/abs-2306-14824}. Another strategy uses LLMs as knowledge experts, combining them with specialists in other fields like vision and speech to tackle diverse tasks~\citep{zeng2023socratic, 2023cafo, liu2023prismer}. Our research concentrates on a third method that leverages the LLM's ability to parse complex queries and delegate tasks to expert agents, whether through custom programs~\citep{Gupta_2023_CVPR}, Python code~\citep{surismenon2023vipergpt}, or dialogue agents~\citep{yang2023mmreact}. However, the effectiveness of these approaches is limited by the quality of the planning sequences and visual experts. 

% Drawing inspiration from the performance improvements seen with step-by-step verification~\citep{Lightman2023LetsVS}, we enhance this line of work by incorporating additional verification strategies.

\paragraph{Compositional multi-modal methods.}
At an early stage, neural module networks (NMN)~\citep{Andreas2015NeuralMN, DBLP:conf/iccv/HuARDS17, Johnson2017InferringAE, DBLP:conf/eccv/HuADS18, DBLP:conf/naacl/LeCH22, DBLP:conf/mm/Qian0DC022,DBLP:conf/aaai/WangWC024} create end-to-end differentiable networks with neural models, but their pre-set modules struggle with open-domain tasks, and the complex embedding and attention mechanisms hinder the creation of intermediate supervision signals.
Recently, the presence of LLMs has made it possible to automatically compose various kinds of finetuned neural models~\citep{zeng2023socratic, Gupta_2023_CVPR, surismenon2023vipergpt, yang2023mmreact, liu2023prismer} or external tools~\citep{Parisi2022TALMTA, DBLP:conf/iclr/KhotTFF0CS23, DBLP:journals/corr/abs-2302-04761, DBLP:journals/corr/abs-2303-17580,DBLP:journals/corr/abs-2304-09842,DBLP:journals/corr/abs-2307-16789}. These works allow us to diagnose the intermedia rationales of the reasoning process. However, human annotation of these intermedia results can be rather time-consuming. In this work, we make ways to correct errors in the intermedia results without any human intervention.

\paragraph{Self-correction in LLMs.}
Although LLMs achieve great success in various tasks, there are many errors in LLM-based system~\citep{Pan2023AutomaticallyCL}: hallucination~\citep{DBLP:journals/corr/abs-2305-11747, Zhang2023HowLM}, unfaithful reasoning~\citep{Golovneva2022ROSCOEAS, Ribeiro2023STREETAM, LYU2023FaithfulCR}, toxic, biased and harmful contents~\citep{Shaikh2022OnST}, flawed code. One way to fix these errors is to use LLMs themselves~\citep{Madaan2023SelfRefineIR,Shinn2023ReflexionLA,selfee2023,Yan2023LearningTS} to obtain feedback to correct the errors. Incorporating self-correction strategies from LLMs, researchers aim to streamline reasoning in multi-modal systems. IPVR~\citep{Chen2023SeeTC} employs LLMs for rationale generation and cross-modality validation to ensure consistent inference. IdeaGPT~\citep{you2023idealgpt} uses an LLM to summarize and iteratively refine the output of visual experts. To overcome the inherent limitations of LLM self-correction, our approach merges LLM feedback with insights from visual experts to authenticate intermediate results and the reasoning process.

\section{\model}
To address the aforementioned shortcomings, we propose \model. This framework adopts exoskeleton verification modules to calibrate the prediction of the execution modules and refine the reasoning path with tree searching. In this section, we will first introduce the preliminaries, including our task definition and visual programming. Then we will show the verification modules, and describe how the verification results are applied to correct the results of execution modules and to search for the reasoning trace. Additionally, we will introduce two mechanisms -- negative sampling and post-hoc self-correction to alleviate extra errors introduced by verification modules.

\subsection{Preliminaries}
\paragraph{Task definition.} Our work focuses on Visual Compositional Reasoning~(VCR) tasks. These VCR tasks require reasoning on a series of steps about an image input $I$ and a text input $T$, and predict the output, \eg answer to a given question, edited images given a language instruction, etc.

\paragraph{Visual programming (\visprog).} 
\visprog~\citep{Gupta_2023_CVPR} is a zero-shot VCR model that leverages LLMs and pretrained vision models. It transforms complex text into a program of operations ($P = \{o^1, \ldots, o^n\}$) using LLMs, which are then executed by various vision models (\eg, object detectors, VQA models). Each operation $o^i$ yields an output $a_i$, where $a_i$ serves as the input for the next operation. The final prediction is made after all operations are executed. However, this approach highlights two key shortcomings of existing approaches: i) module error, the operation models can not predict the answer correctly; ii) planning error, the LLM might generate unfaithful reasoning.

\paragraph{Framework overview}
\cref{fig:framework} depicts the overall framework. For each operation $o^i$, we get a set of candidate answers $\{a^{i}_1, \ldots, a^{i}_{k}\}$, with probabilities $\{p^{i}_1, \ldots, p^{i}_{k}\}$. Unlike \visprog, which directly takes the top answer, we use additional verification modules to verify each candidate answer, thus producing verification scores $\{s^{i}_1, \ldots, s^{i}_{k}\}$. Then we take the verification score $s$ to calibrate the original scores. Additionally, we use the verification scores to search for a program with high verification scores, in order to refine the execution program $P$ by tree-searching.

% The exploration process can be modeled as a tree, where each node is a reason step $e$. At each step, we can obtain $n$ candidate predictions $A=\{a_1, \ldots, a_n\}$ with its prediction scores $S=\{s_1, \ldots, s_n\}$ from the neural model. Our framework aims to select the best prediction from $A$ for each step and get the best reasoning trace from all candidate nodes. 

\subsection{Verification modules}

% \begin{figure}[t!]
%     \centering
%     \includegraphics[width=\linewidth]{figures/neg.pdf}
%     \caption{enhance the verification procedure with negative samples}
%     \label{fig:neg}
% \end{figure}

% \begin{figure}[t!]
%     \centering
%     % \vspace{-1.5em}
%     \includegraphics[width=0.8\linewidth]{figures/ExoViP_verification.png}
%     % \vspace{-0.5em}
%     \caption{Verification Pipeline. Several verifiers collaboratively verify the candidates from operation modules}
%     \label{fig:vefirication}
%     % \vspace{-2.0em}
% \end{figure}

The verification modules aims to verify the candidate answers $\{a^{i}_1, \ldots, a^{i}_{k}\}$ given an operation $o^i$. For example, the \texttt{LOC(nightstand)} operation returns a set of candidate bounding boxes containing a nightstand, then the verification module verifies whether each of the returned boxes contains a nightstand and produces verification scores. 
Our verification module is a mixture of three off-the-shelf sub-verifiers. The output scores of the three verifiers are combined as the final verification score. It is important to emphasize that the verification model does not incorporate any extra pre-trained models; instead, it utilizes the verifiers that are integral to the execution modules of \visprog. To ensure equitable comparisons between modules, we have deliberately chosen these specific sub-verifiers.

\textbf{Image-text matching verifier} calculates the similarity between the whole images and all candidate sentences, which returns the semantic representation of the image-sentence pair. We construct the candidate sentences $\mathcal{T}_{ans}$ by filling the template ``a photo of'' with candidate answers. In this work, we select CLIP~\citep{Radford2021LearningTV} to calculate the similarity between images and sentences, \ie, $s^{itm}_{ans} = \mathrm{ITM}(\mathcal{T}_{ans}, img)$.
% \begin{equation} 
% s^{itm}_{ans} = \mathrm{ITM}(\mathcal{T}_{ans}, img)
% \end{equation}

\textbf{Image captioning verifier} leverages natural language to describe the visual details of the image. We first get the caption of the image $\mathcal{C}_{img}$ by BLIP~\citep{Li2022BLIPBL}.  We then construct the descriptions of candidate answers $\mathcal{C}_{ans}$ with the template "the image describe". Specifically, for candidate question-answer pairs, we initially transform the pair into a sentence before inserting it into the template. After that, we calculate the sentence semantic similarity~\citep{reimers-2019-sentence-bert} between the captions and the constructed descriptions as the verification score, \ie, $s^{cap}_{ans} = sim(\mathcal{C}_{ans},  \mathcal{C}_{img})$ .
% \begin{equation}
% s^{cap}_{ans} = Sim(\mathcal{C}_{ans}, \mathcal{C}_{img})
% \end{equation}

\textbf{Visual question-answering (VQA) verifier} is more flexible than others, which offers us more opportunities to evaluate the advanced relationships between image and language, such as entailment and factual consistency. 
Slightly different from the other two types of models, for the VQA verifier, we design templates \wrt the neural modules. For example, we use ``\texttt{Is there any {object} in the image?}" for the object detection model, and use ``Does this part look like {object} ?" for the classification model used in the abstract reasoning task. We determine the verification score by BLIP~\citep{Li2022BLIPBL} by calculating the difference in answer probabilities $\mathcal{Q}_{ans}$ between ``yes'' and ``no''
 % , \ie, $s^{vqa}_{ans} = \mathrm{VQA}(\mathcal{Q}_{ans}, True) - \mathrm{VQA}(\mathcal{Q}_{ans}, False)$. 
 \begin{equation}
    s^{vqa}_{ans} = \mathrm{VQA}(\mathcal{Q}_{ans}, True) - \mathrm{VQA}(\mathcal{Q}_{ans}, False)
\end{equation}

\paragraph{Verification score} 
After obtaining the scores from each individual verification module, the verification score is averaged over all scores for each given answer, \ie, $s_{ans} = \mathrm{avg}(s^{itm}_{ans}, s^{cap}_{ans}, s^{vqa}_{ans})$
% \begin{equation}
%     s_{ans} = \mathrm{avg}(s^{itm}_{ans}, s^{cap}_{ans}, s^{vqa}_{ans})
% \end{equation}

% \zhuowan{I suggest not highlight this part and therefore here treat it as an additional trick/detail, check if this looks good.} 
\textbf{Negative sampling.}
Empirically, we find that directly applying this verification score does not work well, because the score scales for different kinds of candidates are not well-calibrated. 
% For example, the scores for ``nightstand'' is usually smaller than the scores for ``furniture''.
Motivated by recent works in truthfulness~\citep{li2022mitigating}, commonsense~\citep{Ye2022ImprovingCI}, and bias~\citep{Ruggeri2023AMS}, we propose to take the difference of a candidate answer $a_{j}$ with its semantic opposites $n_{j}$ as the final verification score. More specifically, the semantic opposite $n_{j}$ is selected based on the text embeddings from CLIP~\cite{Radford2021LearningTV}, \ie the word of lowest embedding similarity is selected. For example, the semantic opposite of ``nightstand'' is ``stocking''. We then compute the difference of the verification scores of the candidate answer and its semantic opposites, and get the final verification score. Mathematically, given a candidate answer $a^{j}$, the final verification score is $s_{j} = s_{a_{j}} - s_{n_{j}}$.
% \begin{equation}
%     s_{j} = s_{a_{j}} - s_{n_{j}}
% \end{equation}

\paragraph{Calibration using verification scores}
After obtaining the verification scores of all candidate answers $S = \{s_1, \ldots, s_k\}$, we normalize them as weights and calibrate the candidate predictions, $p^{\prime}_{j} = w_{j} * p_{j}$,
% \begin{equation}
%     p^{\prime}_{j} = w_{j} * p_{j},
% \end{equation}
where $w_j$ is the normalized verification score. More specifically, the verification score $s_j$ is re-scaled to $w_j = \frac{s_j-s_{min}}{s_{max}-s_{min}} \cdot (\tau - \frac{1}{\tau}) + \frac{1}{\tau}$, where $\tau$ is a hyper-parameter controlling the scaling factor $s_{min}, s_{max}$ are the minimum or maximum of all the candidate scores.

% \begin{equation}
%     w_j = \frac{s_j-s_{min}}{s_{max}-s_{min}} \cdot (\tau - \frac{1}{\tau}) + \frac{1}{\tau},
% \end{equation}
% where $\tau$ is the parameter functioned as the scale variable. Finally, we adopt the normalized verification weights to correct the original candidate prediction scores through dot production.

\subsection{Exploration with reasoning trace}

% \begin{figure}[t!]
%     \centering
%     % \vspace{-1.5em}
%     \includegraphics[width=\linewidth]{figures/ExoViP_search.png}
%     % \vspace{-0.5em}
%     \caption{Search of the reasoning trace. We search the reasoning trace through the program tree, based on the verification scores as well as the LLM self-correctness.}
%     \label{fig:search}
%     % \vspace{-2.0em}
% \end{figure}

\begin{wrapfigure}{r}{.5\textwidth}
    \centering
    \vspace{-3.5em}
    \includegraphics[width=\linewidth]{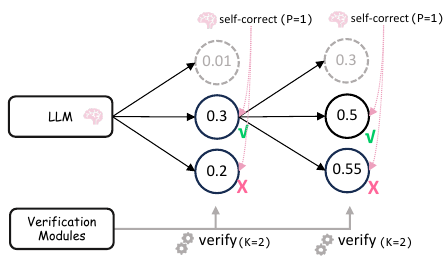}
    % \vspace{-2.5em}
    \caption{Search of the reasoning trace. We beam search through the program tree, based on the verification scores as well as the LLM self-correctness.}
    \label{fig:search}
    \vspace{-1em}
\end{wrapfigure}

To mitigate the planning errors, we further apply the verification scores to refine the reasoning trace predicted by LLMs. Motivated by the recent works showing that searching through a compositional problem space can greatly improve the performance of LLMs for complex tasks~\citep{DBLP:journals/corr/abs-2305-10601, Khalifa2023DiscriminatorGuidedMR, Hao2023ReasoningWL}, we introduce our dynamic reasoning trace searching procedure, which takes advantage of both the LLM self-correctness potential and our verification modules. In \textbf{Appendix~\ref{supp:algo}}, we show the complete algorithm of \model.

\paragraph{Tree-based reasoning trace searching (TRS)} The reasoning trace searching procedure is represented as a tree structure, where each node is a reasoning operation. To get a better reasoning trace, we search from the tree using the beam search algorithm \citep{Graves2012SequenceTW, BoulangerLewandowski2013AudioCR, 43155}, which has long been proven to be effective in sequence-to-sequence problems.

% In each step of searching, we consider both the verification scores and the LLM self-correctness scores. 
% Since the symbolic module is straightforward and less likely to make errors, we only search at each step involving the neural model. 

More specifically, our trace searching procedure contains two steps. First, in order to generate more diverse reasoning traces to search from, we randomly perturb the in-context examples in the prompt for LLM. Second, after we get the result of candidate neural modules, we sort them according to the verification scores and select the top $K$ candidate reasoning traces.

\paragraph{Post-hoc self-correction (PSC)} In some cases, the verification scores can be very close for the top-rated $K$ traces, which could result in suboptimal. Inspired by the zero-shot ranking ability of LLM~\cite{Hou2023LargeLM},  we further use the self-correctness ability of LLMs to rank the $K$ traces and select the top $P$ from them ($P<K$).  More details of the prompts used for LLM self-correction are included in  \cref{supp:prompts}. 
% \zhuowan{More details of the prompts used for LLM self-correction is included in the Appendix XXX. @yuxuan prepare the appendix, also can K and P be shown in Fig-3?.} \yuxuan{update fig3, ExoViP\_searsh\.pptx} 
% If the verification score is zero at some step, we re-plan the search trace.
% The reason is that verification modules can not always make differences between samples that share similar verification scores. 

% We apply this verification strategy repeatedly on the steps requiring neural modules. Compared with existing work \citep{DBLP:journals/corr/abs-2305-10601}, our progressive refining method can improve the planning procedure in a more effective and efficient way. 

\section{Experiments}

% \zzl{Polish and refine. Too many details. Add reference to each subsection.}
In this section, we apply the \model to \visprog, demonstrating the effectiveness of our approach through results and anlysis derived from six distinct tasks, including visual question answering, referring expression understanding, visual reasoning using natural language, abstract reasoning, language-guided image editing, video question answering. Subsequent to this, we delve into an exploration of potential future projections in \cref{exp:discussion}.  For additional information regarding the implementation and experiment setups and selection of baselines, please refer to \textbf{Appendix \ref{supp:implementation}}.

\subsection{Compositional Visual Question Answering}
\label{exp:compositionalvqa}

\subsubsection{Main Results and Analysis}

    % \end{minipage}\hfill

  \begin{wraptable}{r}{.6\linewidth}
    
\resizebox{\linewidth}{!}{
    \begin{tabular}{lp{6.5cm}l}
    \toprule
        & \multicolumn{1}{l}{\bf Methods}  &\multicolumn{1}{c}{\bf Accuracy}
        \\ \midrule
        & BLIP2-xxl~\citep{li2023blip}            & 49.20 \\
        & InstructBLIP-flant5-xl~\citep{Dai2023InstructBLIPTG}             & 55.39 \\ 
        & Llava-1.5-13b*~\citep{DBLP:journals/corr/abs-2310-03744} & 74.56 \\ \midrule
        0 & \visprog~\citep{Gupta_2023_CVPR}         & 57.41 \\ \midrule
        % \visprog + verification         & - \\
        % \visprog + verification & 57.11 \\
        1 & \model w/o self-correctness \& negative sampling \& search & 57.11 \\ 
        % \visprog + verification (counterfactual)         & 58.53 \\
        2 & \model w/o self-correctness \& search & 58.53 \\
        % \visprog + verification (counterfactual) + whole search         & 59.17 \\
        % 3 & \model w/o self-correctness (whole search) & 59.17 \\
        % \visprog + verification (counterfactual) +beam search         & 60.57 \\
        3 & \model w/o self-correctness (TRS) & 60.57 \\
        % \visprog + self-correctness + beam search         & 60.16 \\
        4 & \model w/o verification (PSC) & 60.16 \\
        5  & \model           & 61.49 \\ \bottomrule
    
    \end{tabular}
    }
    \vspace{2mm}
    \captionof{table}{Results of compositional visual question answering on GQA. Llava-1.5-13b* is tuned on GQA training corpora, and evaluated with additional prompt.}
    \vspace{-.1in}
    \label{tab:gqa}
    \end{wraptable}

We evaluate the efficacy of \model on a compositional visual question answering task GQA ~\citep{Hudson2019GQAAN}, and benchmark against top vision-language models, including BLIP2-flant5-xxl~\citep{li2023blip}, InstructBLIP-flan-t5-xl~\citep{Dai2023InstructBLIPTG}, and use LLaVA-1.5-13B~\citep{DBLP:journals/corr/abs-2310-03744} as a reference point. 
% \paragraph{Main Results}  
% We apply our method to \visprog and report the results on GQA in \cref{tab:gqa}. While \visprog has already demonstrated good performance (57.41) compared with BLIP2 and InstructBLIP, our method further improves its performance to 61.49, showing a significant performance boost.
Our method boosts \visprog's score from $57.41$ to $61.49$, surpassing BLIP2 and InstructBLIP, as detailed in \cref{tab:gqa}. It's important to note that our method does not incorporate any additional modules or knowledge compared to \visprog. The verification modules that we use are inherent to \visprog itself.
% To verify the effectiveness of each component in our method, we run a series of analysis experiments on our method (also in \cref{tab:gqa}).
To verify the effectiveness of each component in our method, we run a series of ablation studies on our framework (also in \cref{tab:gqa}).
% The overall results are demonstrated on \cref{tab:analysis}. As we mentioned before, 
We have the following observations:

\paragraph{Negative sampling enhances the robustness of \model}
Merely adding verification modules to a system (Line-1) is not sufficient for achieving better results; it may actually lead to decreased performance. On the other hand, when we implement a negative sampling technique that utilizes semantic opposites in these verification modules (Line-2), there is a marked enhancement in the system's performance. We posit that this approach could be instrumental in reducing the likelihood of new errors being introduced.

% In line-3, ``whole search" means we use LLMs to obtain a set of complete planning traces, then execute all the traces to get the final verification scores, and select the best trace with the highest verification score. 
% The Tree-based Reasoning Trace Searching (TRS) strategy employs a methodical approach to decision-making by selecting the next step based on the highest verification scores, thereby optimizing the use of these scores. Empirical evidence shows that implementing TRS can significantly improve accuracy, increasing it from an initial $58.53$ to $60.57$. This marked improvement in precision underscores the effectiveness of our verification-based search strategy, which has the potential to resolve numerous planning errors. 
\paragraph{TRS effectively utilizes the verification score} Empirical evidence shows that implementing TRS can increase accuracy from an initial $58.53$ to a subsequent $60.57$ (Line-3). This improvement in precision underscores the effectiveness of our verification-based search strategy, which has the potential to resolve numerous planning errors.

\paragraph{Verification mechanism enhance LLM self-correction} In Line-4, we exclusively utilize the post-hoc self-correction (PSC) of the LLM during the process of trace searching, eschewing the use of verification scores. The findings demonstrate an enhancement in accuracy of $2.75$ when compared to the original \visprog. However, the implementation of both verification scores and LLM self-correctness concurrently results in a superior performance enhancement.
% \item \textit{\model achieves the best performance with the collaboration of all the introduced components.} In line-6, we combine all the introduced components and get the final performance of $61.49$, showing a significant boost ((4.08). We believe our method successfully incorporates the verification modules with the LLM self-correctness ability. 

% \end{enumerate}

% \begin{table}[]
%     \centering
%     \captionof{table}{Analysis on the sub-verifiers.}
%     \label{tab:analysis-verify}
%     \begin{tabular}{lc}
%          \multicolumn{1}{l}{\bf Methods}  &\multicolumn{1}{c}{\bf Accuracy} \\
%         \midrule
%         Base             & 58.14 \\ \midrule
%         ITM             & 59.26 \\
%         Caption         &59.22 \\ 
%         VQA             &59.35 \\ \midrule
%         All & 60.03 \\ \bottomrule
%     \end{tabular}
%     % \caption{Caption}
%     % \label{tab:my_label}
% \end{table}

% \begin{figure*}[t]
%     \centering
%     % \vspace{-0.5em}
%     \includegraphics[width=0.98\linewidth]{figures/exovip_case_imageediting.pdf}
%     % \vspace{-0.5em}
%     \caption{Qualitative results of text-guided image editing on MagicBrush.}
%     \label{fig:magic-brush-qualitative}
%     % \vspace{-1.0em}
% \end{figure*}

\subsubsection{Analysis Experiments}

We experiment to explore how the tree-based search algorithm and verification scores interact to improve our approach. The algorithm uses scores to guide branching, while insights from the search enhance score contrasts, refining path differentiation and aiding in finding the best solution. All experiment settings are aligned with the Appendix ~\ref{supp:proof}.

% \begin{table}[t!]
%     % \vspace{-.5em}
%     \centering
    
%     \begin{center}
%     % \resizebox{\linewidth}{!}{
%     \begin{tabular}{lc} \toprule
%          \multicolumn{1}{l}{\bf Methods}  &\multicolumn{1}{c}{\bf Accuracy} \\
%         \midrule
%         Base             & 58.14 \\ \midrule
%         Image-text matching verifier             & 59.26 \\
%         Image caption verifier         &59.22 \\ 
%         Visual question answering verifier             &59.35 \\ \midrule
%         All & 60.03 \\ \bottomrule
%     \end{tabular}
%     % }
%     \vspace{-1em}
%     \end{center}
%     \captionof{table}{Analysis on the sub-verifiers.}
%     \vspace{-1em}
%     \label{tab:analysis-verify}
% \end{table}

% \begin{figure}[t!]
%     \vspace{-0.5em}
%     \centering
% 	\includegraphics[width=0.9\linewidth]{figures/dist.pdf}
%     \vspace{-1.0em}
%     \caption{Distribution of verification scores w. and w/o trace searching.}
%     % \vspace{-0.5em}
%     \label{fig:dist}
% \end{figure}

    \begin{wraptable}{r}{.4\linewidth}
 %    \vspace{-.5em}
 %    \centering
	% \includegraphics[width=\linewidth]{figures/dist.pdf}
 %    \vspace{-0.9em}
 %    \captionof{figure}{Distribution of verification scores w. and w/o trace searching. }
 %    \vspace{-0.5em}
 %    \label{fig:dist}   
    \begin{tabular}{lc} \toprule
         \multicolumn{1}{l}{\bf Methods}  &\multicolumn{1}{c}{\bf Accuracy} \\
        \midrule
        Base             & 58.14 \\ \midrule
        Image-text Matching             & 59.26 \\
        Image Caption         &59.22 \\ 
        Visual QA             &59.35 \\ \midrule
        All & 60.03 \\ \bottomrule
    \end{tabular}
        % \vspace{-0.5em}
        \vspace{2mm}
    \captionof{table}{Analysis on the sub-verifiers.}
    \vspace{-.1em}
        % \vspace{-0.4em}
    \label{tab:analysis-verify}

    % \captionof{table}{Analysis on the sub-verifiers.}
    % \label{tab:analysis-verify}
    % % \begin{center}
    % % \resizebox{\linewidth}{!}{
    % \begin{tabular}{lc} \toprule
    %      \multicolumn{1}{l}{\bf Methods}  &\multicolumn{1}{c}{\bf Accuracy} \\
    %     \midrule
    %     Base             & 58.14 \\ \midrule
    %     ITM             & 59.26 \\
    %     Caption         &59.22 \\ 
    %     VQA             &59.35 \\ \midrule
    %     All & 60.03 \\ \bottomrule
    % \end{tabular}
    % % }
    \vspace{-1em}
    % \end{center}
    \end{wraptable}

\paragraph{Mixture of Sub-verifers.}
We evaluate the effects of different types of verification modules with the setting of the best demonstration setting. As is illustrated in \cref{tab:analysis-verify}, Different verification modules share similar boost gain, but a mixture of these modules can benefit more.

% \begin{wrapfigure}{r}{.4\textwidth}
%     \centering
%     \vspace{-0.5em}
%     \includegraphics[width=\linewidth]{figures/dist.pdf}
%     \vspace{-0.5em}
%     \caption{Distribution of verification scores w. and w/o trace searching.}
%     \label{fig:dist}
% \end{wrapfigure}

% \begin{table}[t!]
    
%     % \vspace{-0.5em}
%     \centering
%     \begin{center}
%     \begin{tabular}{lc}
%     \toprule
%         \multicolumn{1}{l}{\bf Methods}  &\multicolumn{1}{c}{\bf Avg. Steps}
%         \\ \midrule
%         \visprog~\citep{Gupta_2023_CVPR}         &5.77 \\
%         \model             & 4.92 \\ \bottomrule
%     \end{tabular}
%     \end{center}
%     \captionof{table}{Comparisons on planning steps.}
%     \label{tab:SPL}
% \end{table}

% \paragraph{Collaboration of TRS and Sub-Verifiers.}We have computed the verification scores across various samples and depicted their distribution in \cref{fig:dist}. Our analysis reveals two significant improvements brought about by the implementation of our trace-searching strategy. Firstly, the average verification score experienced a substantial increase following the application of our trace-searching strategy. This suggests that our method has effectively enhanced the performance of the system. Secondly, we observed a greater variance in the verification scores post the application of our method. This indicates that our approach can potentially leverage these verification scores to enhance the efficacy of the reasoning traces. 
\paragraph{Enhanced Verification through TRS.} Our analysis, illustrated in \cref{fig:dist}, shows that implementing our trace-searching strategy significantly improved verification scores. Additionally, the increased variance in scores suggests our method could further refine the effectiveness of reasoning traces.

\paragraph{Enhanced Planning Efficacy}
In our comparative analysis of the GQA task, we observed that our method, \model, significantly outperforms the baseline, \visprog, in terms of planning efficiency and accuracy. The average number of planning steps required by \model decreased from 5.92 to 4.77, indicating that the TRS strategy employed by our method streamlines the planning process, allowing for a more direct path to the final plan. We also compute the average inference time, which is shown on \textbf{Appendix~\ref{supp:efficiency}}. Furthermore, we noted a reduction in the error rate, with the percentage of unexecutable plans dropping from $5.84\%$ to $3.82\%$. This demonstrates that \model not only reduces the complexity of the planning process but also enhances the reliability of the generated plans, predicting a higher number of executable routines compared to the baseline.

\subsection{Abstract Visual Reasoning}
\label{exp:abstract-reasoning}
% \begin{table}[t!]
    
%     % \vspace{-0.5em}
%     \centering
%     \begin{center}
%     \resizebox{0.8\linewidth}{!}{
%     \begin{tabular}{lc} \toprule
%         \multicolumn{1}{l}{\bf Methods}  &\multicolumn{1}{c}{\bf Accuracy}
%         \\ \midrule
%         CLIP-large~\citep{Radford2021LearningTV}             & 27.26 \\
%         \model w/o verification         & 24.46 \\
%         \model             &26.22 \\ \bottomrule
%     \end{tabular}}
%     \end{center}
%     \captionof{table}{Results of abstract reasoning on KILOGRAM.}
%     \label{tab:kilogram}
% \end{table}

% Details of setups and selections of baselines are listed at ~\ref{supp-detail:kilogram}

% \begin{figure}[th!]
%     \centering
%     \includegraphics[width=\linewidth]{figures/cases/exovip_qualitative_kilogram.pdf}
%     \caption{Implementation of \model on abstract reasoning.}
%     \label{fig:kilogram}
% \end{figure}

% \begin{wrapfigure}{r}{.45\textwidth}
%     \centering
%     \includegraphics[width=\linewidth]{figures/cases/exovip_qualitative_kilogram.pdf}
%     \caption{Implementation of \model on abstract reasoning.}
%     \label{fig:kilogram}
% \end{wrapfigure}
We tested model performance on the KILOGRAM~\citep{Ji2022AbstractVR} dataset's text-to-image retrieval task, involving 1,251 tangram puzzles with abstract shape recognition. Accuracy was the main evaluation metric. The CLIP-large model~\citep{Radford2021LearningTV} served as our baseline for the text-to-image retrieval task.
As is illustrated in ~\cref{fig:kilogram}, in our approach, we leverage the LLM to identify potential semantic components of a given description. Simultaneously, we segment the image into distinct visual parts. Following this, we align the identified semantic parts with their corresponding visual counterparts to optimize the matching process. \cref{tab:kilogram} illustrates how our method effectively utilizes conceptual components to enhance abstract image understanding. However, a performance gap is still noticeable when compared to CLIP. Despite our method narrowing this gap, it is still unable to reach SOTA performance levels. The importance of part identification in human abstraction has been well-established in prior research \citep{Tversky1984ObjectsPA}. We posit that the efficacy of our approach could be significantly improved by integrating a more advanced scene segmentation model.
% For our method, given an object, we adopt the LLM to get its possible semantic parts. At the same time, we segment the image into several visual parts. After that, we align the semantic parts with the visual parts to enhance the matching process.   In \cref{tab:kilogram}, we find our method could take advantage of concept parts to improve abstract image understanding. However, we still see the gap between \visprog and CLIP. Although our method decreases the performance gap, the compositional method still can not achieve SOTA. Since part identification has already been demonstrated to play an important role in human abstraction~\cite{Tversky1984ObjectsPA}. We believe our method can be enhanced by introducing a better scene segmentation model.

\begin{table}[t!]
\begin{minipage}[t]{.48\textwidth}
    
    \begin{center}
    \captionof{table}{Visual referring expression on RefCOCO, RefCOCO+, and RefCOCOg.}
        \vspace{-0.5em}
    \resizebox{\linewidth}{!}{
    \begin{tabular}{lc}
    \toprule
        \multicolumn{1}{l}{\bf Methods}  &\multicolumn{1}{c}{\bf IoU}
        \\ \midrule
        Qwen-vl-chat-7b~\citep{Bai2023QwenVLAV}             & 32.54 \\
        \visprog~\citep{Gupta_2023_CVPR}         &27.28 \\
        \model             &31.50 \\ \bottomrule
    \end{tabular}
    }
    \vspace{.5em}
    \label{tab:refcoco}
    \end{center}
\end{minipage}\hfill
\begin{minipage}[t]{.48\textwidth}
    
    % \vspace{-0.5em}
    \captionof{table}{Abstract reasoning on KILOGRAM.}
            \vspace{-0.2em}
    \begin{center}
    \resizebox{\linewidth}{!}{
    \begin{tabular}{lc} \toprule
        \multicolumn{1}{l}{\bf Methods}  &\multicolumn{1}{c}{\bf Accuracy}
        \\ \midrule
        CLIP-large~\citep{Radford2021LearningTV}             & 27.26 \\
        \visprog~\citep{Gupta_2023_CVPR}         & 24.46 \\
        \model             &26.22 \\ \bottomrule
    \end{tabular}
    }
    \vspace{.5em}
    \label{tab:kilogram}
    \end{center}
\end{minipage}\\

\begin{minipage}[t]{.46\textwidth}
    
    \begin{center}
    \captionof{table}{Visual reasoning on NLVR2.}
            \vspace{-0.2em}

    \resizebox{\linewidth}{!}{
    \begin{tabular}{lc}
    \toprule
        \multicolumn{1}{l}{\bf Methods}  &\multicolumn{1}{c}{\bf Accuracy}
        \\ \midrule
        OFA-large~\citep{Wang2022UnifyingAT}             &58.38 \\
        \textsc{\visprog}~\citep{Gupta_2023_CVPR}         &67.66 \\
        \model             &67.96 \\ \bottomrule
    \end{tabular}
    }
    % \vspace{-0.5em}
    \label{tab:nlvr}
    \end{center}
    \vspace{-.2em}
\end{minipage}\hfill
% \end{table}
\begin{minipage}[t]{.5\textwidth}
    
    % \vspace{-0.5em}
    \begin{center}
    \captionof{table}{Image editing on MagicBrush.}
            \vspace{-0.5em}

    \resizebox{\linewidth}{!}{
    \begin{tabular}{lcc} \toprule
        \multicolumn{1}{l}{\bf Methods}  & \multicolumn{1}{c}{\bf CLIP-I} &\multicolumn{1}{c}{\bf DINO}
        \\ \midrule
        InstructPix2Pix~\citep{Brooks2022InstructPix2PixLT}              & 84.19 & 69.60 \\
        \visprog~\citep{Gupta_2023_CVPR}    &90.82 &82.70 \\
        \model              &91.27 &83.40\\ \bottomrule
    \end{tabular}
    }
    \label{tab:magicbrush}
    \end{center}
    \vspace{-.2em}
\end{minipage}
\end{table}

\begin{figure}[t!]
    \centering
    \includegraphics[width=.9\linewidth]{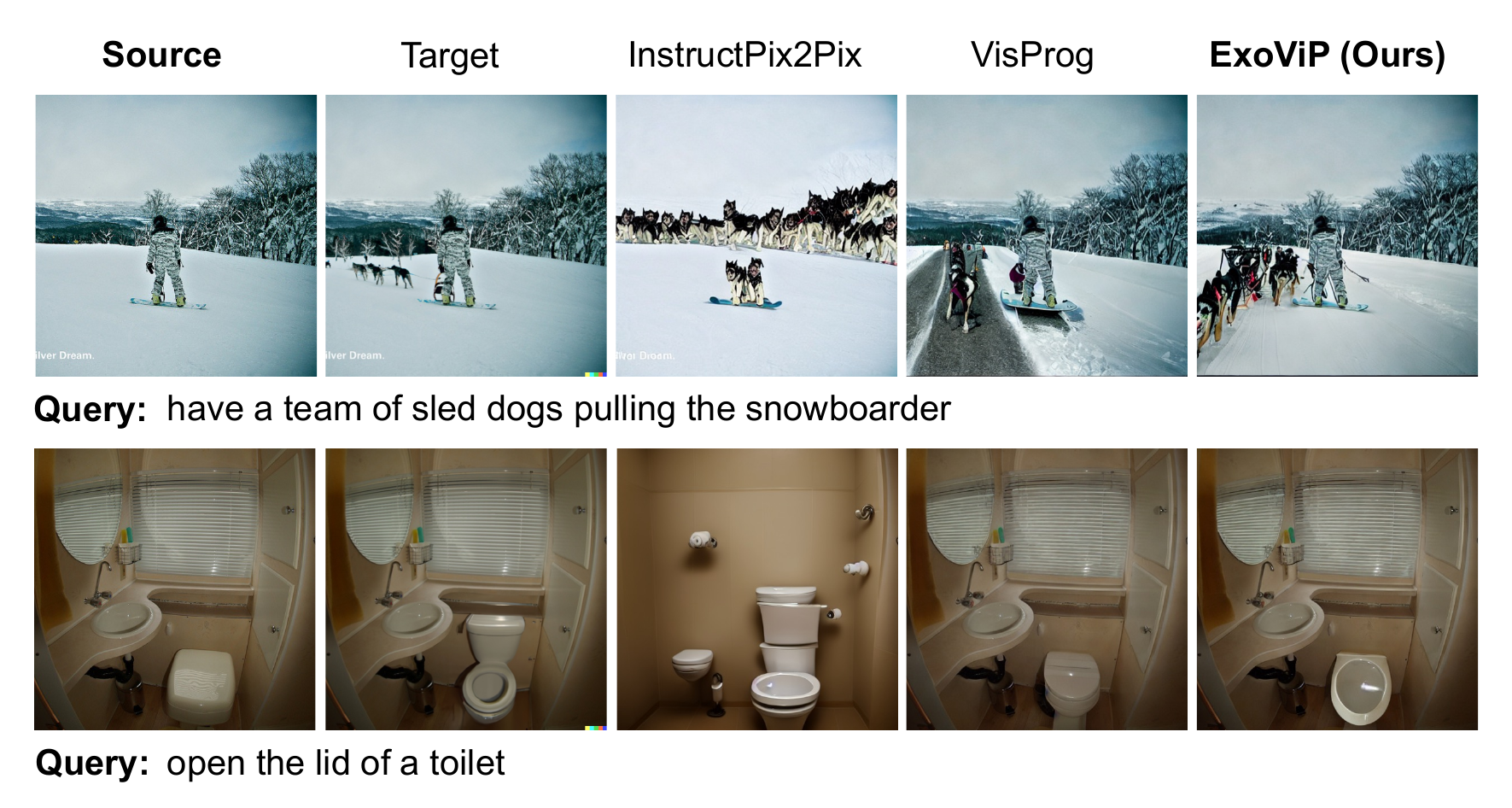}
    \vspace{-.05in}
    \caption{Qualitative results of text-guided image editing on MagicBrush}
    % \vspace{-.1in}
    \label{fig:case_imageeditting}
    \vspace{-.2in}
\end{figure}

\subsection{Language-grounded Visual Tasks}
\label{exp:language-grounded}

% \paragraph{Setup} Given a natural language query describing a region in a given image, the visual language grounding task requires identifying the bounding box of the object in the image being referred to. RefCOCO, RefCOCO+ and RefCOCOg~\citep{2016refcoco, kazemzadeh-etal-2014-referitgame} are three standard datasets for this task. For our study, we have randomly selected two samples per object type from the test sets of both the RefCOCO, RefCOCO+ and RefCOCOg datasets. Consequently, our test set comprises a total of 418 queries. We evaluate all the results by intersection-over-union~(IoU).
% % We randomly sample 2 samples per type from the test set from the RefCOCO dataset and RefCOCO+ dataset. In summary, our test set includes 66 types, \eg ``bicycle'', ``backpack'', and 261 queries. We evaluate all the results by intersection-over-union~(IoU).

% \paragraph{Baseline model} We adopt the \code{Qwen-vl-chat-7b}~\citep{Bai2023QwenVLAV}, a pre-trained large vision-language model that uses Qwen-7B with further training with aligned techniques, as the baseline. \code{Qwen-vl} outperforms current SOTA generalist models on multiple VL tasks and has a comprehensive coverage in terms of capability range.

\paragraph{Visual Referring Expressions~(VRE)}
\label{exp:visual-referring-expression}

% \begin{table}[t!]
    
%     % \vspace{-0.5em}
%     \centering
%     \begin{center}
%     \resizebox{0.8\linewidth}{!}{
%     \begin{tabular}{lc}
%     \toprule
%         \multicolumn{1}{l}{\bf Methods}  &\multicolumn{1}{c}{\bf IoU}
%         \\ \midrule
%         Qwen-vl-chat-7b~\citep{Bai2023QwenVLAV}             & 27.28 \\
%         \visprog~\citep{Gupta_2023_CVPR}         &22.60 \\
%         \model             &25.61 \\ \bottomrule
%     \end{tabular}}
%     \end{center}
%     \captionof{table}{Results of visual referring expression on subset of RefCOCO, RefCOCO+ and RefCOCOg.}
%     \label{tab:refcoco}
% \end{table}

Our study on VRE used a subset of the RefCOCO, RefCOCO+, and RefCOCOg datasets~\citep{2016refcoco, kazemzadeh-etal-2014-referitgame}, and evaluated using intersection-over-union (IoU). We benchmarked against the \code{Qwen-vl-chat-7b}~\citep{Bai2023QwenVLAV} model, a high-performing, pre-trained vision-language model. 
The results presented in \cref{tab:refcoco} illustrate that, even though our approach does not reach the state-of-the-art (SOTA) performance achieved by \code{Qwen-vl} on the RefCOCO dataset, it nonetheless narrows the gap between \visprog and large vision-language models. \code{Qwen-vl} is a highly complex model, constructed on a language learning model (LLM) consisting of 7 billion parameters and trained on a corpus of trillions of tokens. In contrast, our approach utilizes a team of specialized experts, whose combined parameters amount to less than 1 billion. We are optimistic that the performance of our method can be further enhanced by incorporating more sophisticated experts.
% As demonstrated in \cref{tab:refcoco}, although our method can't achieve SOTA (\code{Qwen-vl}) on the RefCOCO dataset, it helps bridge the gap between \visprog and the large vison-language model. While \code{Qwen-vl} is built on an LLM with 7 billion parameters and trained on trillions of tokens from the corpus, our method assembles a team of experts whose collective parameters total less than 1 billion. We believe our method can be improved with more advanced experts. 

% \paragraph{Setup} In NLVR2~\citep{suhr-etal-2019-corpus}, given a description of a collection of images, the model needs to justify whether the description is correct or not (binary classification). The task requires dealing with various kinds of linguistic phenomena, like numerical expressions, quantifiers, coreference, negation, \etc. In this work, we use the NLVR2 balanced test set for evaluation, which includes 2,316 questions and corresponding image pairs. To assess the performance of various models, we have chosen accuracy as our primary evaluation metric.

% \paragraph{Baseline model} We consider the OFA-large~\citep{Wang2022UnifyingAT} as our baseline model. OFA offers a unified approach to handle various cross-modal and unimodal tasks within a straightforward sequence-to-sequence learning framework. Since NLVR involves two images, many MLLM models are unable to handle this task. Therefore, we choose OFA-large as our reference model.

\paragraph{Natural Language Visual Reasoning}
\label{exp:nlvr}

% \begin{table}[t!]
    
%     \centering
%     \begin{center}
%     \resizebox{0.8\linewidth}{!}{
%     \begin{tabular}{lc}
%     \toprule
%         \multicolumn{1}{l}{\bf Methods}  &\multicolumn{1}{c}{\bf Accuracy}
%         \\ \midrule
%         OFA-large~\citep{Wang2022UnifyingAT}             &58.38 \\
%         \textsc{\visprog}~\citep{Gupta_2023_CVPR}         &67.66 \\
%         \model             &67.96 \\ \bottomrule
%     \end{tabular}}
%     \end{center}
%     \captionof{table}{Results of visual reasoning on NLVR.}
%     % \vspace{-0.5em}
%     \label{tab:nlvr}
% \end{table}

In this work, we use the NLVR2~\citep{suhr-etal-2019-corpus} balanced test set for evaluation and use the OFA-large~\citep{Wang2022UnifyingAT} as our baseline, unlike many multi-modal language models that struggle with dual-image inputs. 
\cref{tab:nlvr} presents our findings. While \visprog demonstrates a strong capability for complex reasoning compared to the end-to-end model, our method struggles to enhance its performance significantly. We attribute this to our sole reliance on VQA modules for solving NLVR problems.  Specifically, the efficacy of the decomposed VQA steps is intrinsically constrained by the performance of the VQA model itself. This limitation becomes especially troublesome when errors accumulate over a series of VQA steps, consequently hampering the overall performance. As a path forward, we foresee potential advancements in the planning process, which could involve the integration of a wider array of expert inputs.

\paragraph{Text-guided Image Editing}
\label{exp:image-editing}

% \begin{table}[t!]
    
%     \centering
%     \begin{center}
%     \resizebox{0.8\linewidth}{!}{
%     \begin{tabular}{lcc} \toprule
%         \multicolumn{1}{l}{\bf Methods}  & \multicolumn{1}{c}{\bf CLIP-I} &\multicolumn{1}{c}{\bf DINO}
%         \\ \midrule
%         InstructPix2Pix~\citep{Brooks2022InstructPix2PixLT}              & 84.19 & 69.60 \\
%         \visprog~\citep{Gupta_2023_CVPR}    &90.82 &82.70 \\
%         \model              &91.27 &83.40\\ \bottomrule
%     \end{tabular}
%     }
%     \end{center}
%     \captionof{table}{Results of image editing on MagicBrush.}
%     \vspace{-1.0em}
%     \label{tab:magicbrush}
% \end{table}

We use the MagicBrush dataset~\citep{Zhang2023MagicBrushAM}for evaluation. Image quality is gauged using CLIP-I and DINO embeddings for similarity assessment. Our baseline is the GPT3-augmented InstructPix2Pix~\citep{Brooks2022InstructPix2PixLT} model.
The results from both CLIP-I and DINO are presented in~\cref{tab:magicbrush}. These results illustrate the capability of our method to enhance the similarity between the edited image and the target image, signifying the precision of our image editing technique. For a more comprehensive evaluation of the editing quality, we have conducted a case study. ~\cref{fig:case_imageeditting} exhibits some instances using MagicBrush. It is observed that non-compositional methods, \ie InstructPix2Pix, tend to alter unrelated pixels, whereas compositional methods, \ie \visprog and our model, offer more control. 
Furthermore, when compared to \visprog, our method excels in two key areas: accurately pinpointing the region that requires editing, and adjusting the image to the appropriate extent. This demonstrates the superiority of our method in both localization and modification of the image.

\subsection{Spatial-Temporal Video Reasoning}
\label{exp:spatial-temporal}

\begin{wraptable}{r}{.5\textwidth}
    \centering
    \begin{center}
    \resizebox{0.95\linewidth}{!}{
    \begin{tabular}{lc} \toprule
        \multicolumn{1}{l}{\bf Methods}  & \multicolumn{1}{c}{\bf Accuracy}
        \\ \midrule
        Video-LLaVA~\citep{DBLP:journals/corr/abs-2311-10122}              & 30.38  \\
        \model w/o verification    & 37.88  \\
        \model              & 38.00 \\ \bottomrule
    \end{tabular}
    }
    \end{center}
    % \vspace{-1.0em}
    \captionof{table}{Results of Spatial-Temporal Reasoning on AGQA.}
    \vspace{-1.0em}
    \label{tab:agqa}
\end{wraptable}

% \begin{table}[t!]
    
%     \centering
%     \begin{center}
%     \resizebox{0.8\linewidth}{!}{
%     \begin{tabular}{lc} \toprule
%         \multicolumn{1}{l}{\bf Methods}  & \multicolumn{1}{c}{\bf Accuracy}
%         \\ \midrule
%         Video-LLaVA~\citep{DBLP:journals/corr/abs-2311-10122}              & 30.38  \\
%         \model w/o verification    & 37.88  \\
%         \model              & 38.00 \\ \bottomrule
%     \end{tabular}
%     }
%     \end{center}
%     \vspace{-1.0em}
%     \captionof{table}{Results of Spatial-Temporal Reasoning on AGQA.}
%     \vspace{-1.0em}
%     \label{tab:agqa}
% \end{table}

% \paragraph{Setup} AGQA~\citep{DBLP:conf/cvpr/Grunde-McLaughlin21} is a dataset designed for video-based question-answering tasks that require compositional spatio-temporal reasoning. For our experiments, we utilized AGQA 2.0~\citep{DBLP:journals/corr/abs-2204-06105} as the benchmark and randomly selected 100 question-answer pairs across various reasoning categories, yielding a total of 800 unique pairs for analysis. We assessed the performance of our approach using accuracy as the primary evaluation metric.

% \paragraph{Baseline model} We utilize Video-LLaVA~\citep{DBLP:journals/corr/abs-2311-10122} as our benchmark reference. Video-LLaVA represents a state-of-the-art, expansive vision-language model that is trained on a diverse dataset comprising both images and videos. It has demonstrated exceptional performance, setting new standards across a wide spectrum of image and video benchmarks.

% \paragraph{Setup and Baselines}
We conducted experiments using the subset of AGQA 2.0 dataset~\citep{DBLP:journals/corr/abs-2204-06105}. Our reference was the Video-LLaVA~\citep{DBLP:journals/corr/abs-2311-10122}, a top-tier vision-language model known for its superior performance on numerous benchmarks.
In our methodology, we address the question by breaking it down into temporal and spatial components. For the temporal aspect, we aim to find the event or action within a video. This is achieved by uniformly sampling frames from the video and generating corresponding captions. We then compute the sentence similarity between these captions and the input query. Subsequently, we identify the event by locating the video segment with the highest similarity, utilizing a monotonic stack algorithm. By adopting this approach, we can effectively mitigate the Out-of-Vocabulary (OOV) issue that plagues current action classification models. Regarding the spatial component, it is predominantly addressed by existing VQA models. The experimental outcomes, as presented in ~\cref{tab:agqa}, indicate that the compositional method yields strong performance. However, the benefits brought by verification are limited. Upon further examination of the results and the underlying reasoning paths, we observe that the majority of the unsuccessful cases can be attributed to the performance of the VQA models, a trend that aligns with findings from the NLVR task.

\subsection{Discussion}
\label{exp:discussion}

\subsubsection{Error analysis of \visprog and \model}
\label{supp:failure-case}

% \begin{figure}[h]
%     \centering
%     \includegraphics[width=.3\linewidth]{figures/error.pdf}
%     \caption{Distribution of the failure cases of original \visprog.}
%     \label{fig:error}
% \end{figure}

% \begin{figure}[th!]
%     % \vspace{-0.5em}
%     \centering
%     \begin{minipage}{.45\textwidth}
%         \centering
% 	\includegraphics[width=0.9\linewidth]{figures/error_before.pdf}
%     \vspace{-0.9em}
%     \captionof{figure}{Distribution of the failure cases of original \visprog.}
%     \vspace{-0.5em}
%     \label{fig:error-before}
%     \end{minipage}\hfill
%     \begin{minipage}{.45\textwidth}
%         \centering
% 	\includegraphics[width=\linewidth]{figures/error_after.pdf}
%     \vspace{-0.9em}
%     \captionof{figure}{Distribution of the failure cases of \model.}
%     \vspace{-0.5em}
%     \label{fig:error-after}
%     \end{minipage}
% \end{figure}

\begin{figure}
    \centering
    \includegraphics[width=0.7\linewidth]{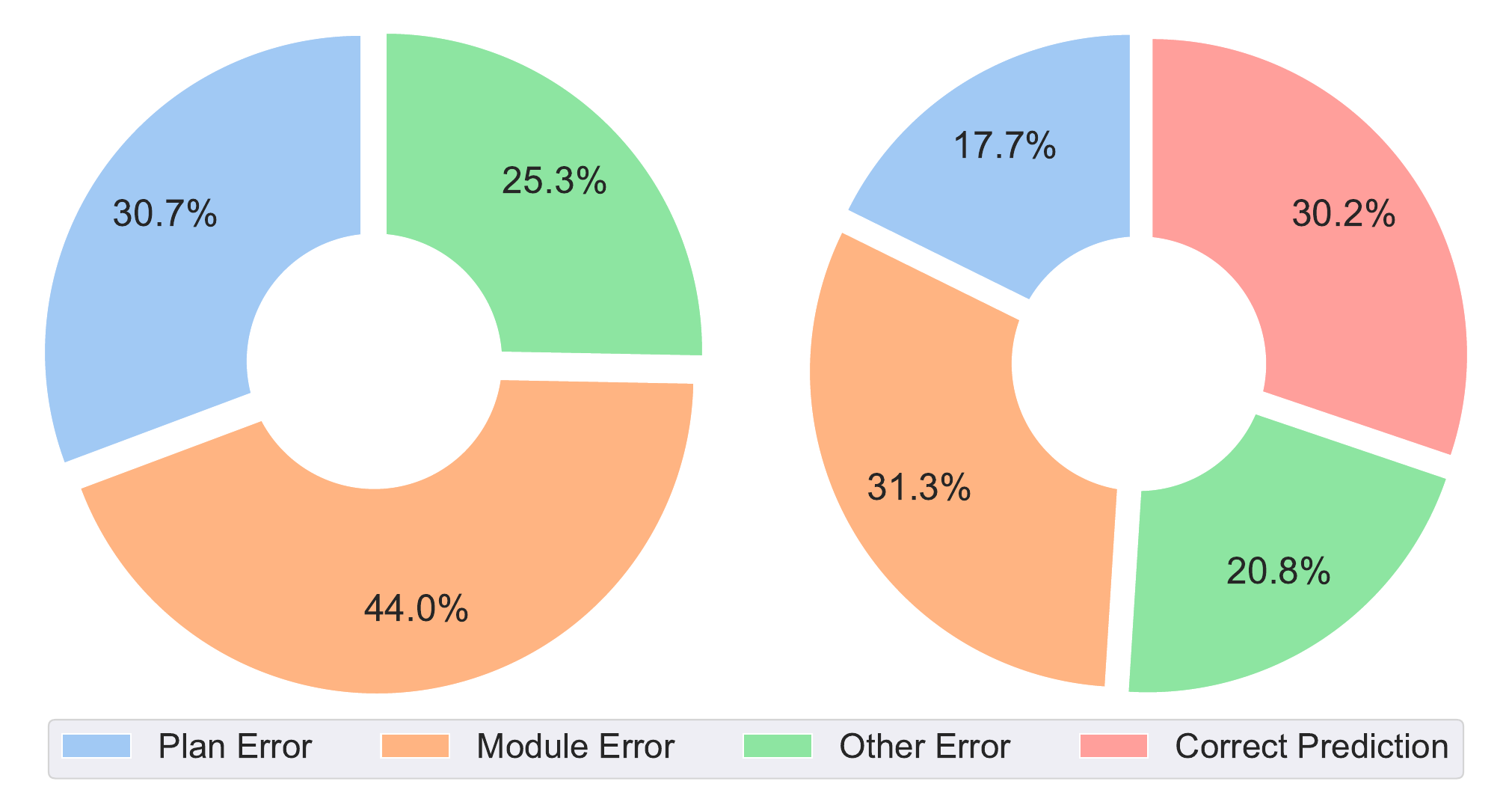}
    \caption{Distribution of the failure cases of original \visprog (left), and distribution of the failure cases of \model (right)}
    \label{fig:error-analysis}
    \vspace{-.2in}
\end{figure}

We manually analyze 100 randomly sampled failure cases on \visprog. We find that there are three typical reasons for the failures: (a) vision module prediction error; (b) LLM planning error; (c) others. We demonstrate the statistics of the failure cases in~\cref{fig:error-analysis} (left). Following the application of our proposed framework, we reassessed the same cases in ~\cref{fig:fig:error-analysis} (right) and were pleased to discover a reduction in module errors by $28.87\%$, and a decrease in planning errors by $42.35\%$. Nevertheless, our current strategy was unable to rectify $69.8\%$ of the errors. When juxtaposed with the data from \cref{tab:gqa}, our method has enhanced \visprog by $7.11\%$, which is lower than the improvement of the failure cases. This outcome suggests that our approach may give rise to novel challenges. We further demonstrate common errors of our method in \cref{fig:failurecase-module} and \cref{fig:failurecase-plan}. We find the majority of these failure cases originate from the VQA module.

\paragraph{Additional error analysis}
Our methodology acknowledges the possibility that the inclusion of verifiers might inadvertently increase the error rate. To counteract this, we have adopted a negative sampling approach, and we have integrated with a verification score and the inherent self-corrective feature of LLMs. The efficacy of these combined strategies in reducing the incidence of additional errors is evidenced by the results displayed in ~\cref{tab:analysis-verify}. Nonetheless, while our approach successfully diminishes errors related to planning and module execution, it can occasionally lead to the introduction of new errors. Moving forward, we aim to enhance our system by incorporating a greater number of verifiers to more effectively resolve these issues.

% \paragraph{Role of verification modules} In this study, we selected three types of vision-language models as verification modules: Image-text matching verifier, Image captioning verifier, and Visual question-answering verifier. These modules evaluate the alignment between visual and linguistic contents at varying granularities. It is crucial to note that these verification modules, all neural models trained on extensive corpora, are constrained by their respective types. Looking forward, we anticipate a more diverse pool of verifiers not restricted to a specific group of experts. Moreover, the process of selecting verifiers could eventually be automated by the LLMs.
% In this work, we choose three types of vision-language models as verification modules: Image-text matching verifier, Image captioning verifier, and Visual question-answering verifier. These modules assess the alignment between vision and language contents through different granularities. It's important to acknowledge that the verification modules are constrained by their respective types, all of which are neural models trained on large corpora. Looking ahead, we anticipate that the pool of verifiers will not be confined to a specific group of experts. Furthermore, the process of selecting verifiers could potentially be automated by the LLMs.

\subsubsection{Method generalizability}
% \vspace{-.1in}

\begin{wraptable}{r}{.5\textwidth}
    \vspace{-0.5em}
    
    % \vspace{-0.5em}
    \begin{center}
    \resizebox{\linewidth}{!}{
    \begin{tabular}{ll} \toprule
        \multicolumn{1}{c}{\bf Methods}  &\multicolumn{1}{c}{\bf Accuracy}
        \\ \midrule
        ViperGPT~\citep{surismenon2023vipergpt}         & 45.47 \\
        ViperGPT+ExoViP             & 46.84 \\ \bottomrule
    \end{tabular}
    }
    % \vspace{-0.5em}
    \end{center}
    \caption{Results for ViperGPT on GQA.}
        \vspace{-.2em}
    \label{tab:viper}
\end{wraptable}

% \begin{table}[t!]
    
%     \centering
%     \begin{center}
%     \resizebox{0.8\linewidth}{!}{
%     \begin{tabular}{ll} \toprule
%         \multicolumn{1}{c}{\bf Methods}  &\multicolumn{1}{c}{\bf Accuracy}
%         \\ \midrule
%         ViperGPT~\citep{surismenon2023vipergpt}         & 45.47 \\
%         ViperGPT+ExoViP             & 46.84 \\ \bottomrule
%     \end{tabular}}
%     \end{center}
%     \vspace{-1.0em}
%     \caption{Results for ViperGPT on GQA.}
%     \vspace{-1.0em}
%     \label{tab:viper}
% \end{table}

To validate the generalizability of our method, we applied it to  ViperGPT, which composes available modules by generating Python codes. We equip ViperGPT with our method and test its performance on the GQA dataset. The results, presented in \cref{tab:viper}, reveal a less significant performance boost compared to \visprog. We attribute this to ViperGPT providing only a few demonstration examples and adjusting the parameters of the code-generation model to deterministically generate subroutines. We believe this could be improved by introducing diverse demonstrations, similar to \visprog.
% We analyze this due to ViperGPT provides a few examples in the demonstration and it turns the parameter of the code-generation model to make it deterministic to generate subroutines. In other words, ViperGPT benefits little from our reasoning trace-searching strategy.

\section{Conclusion}
% Current problems determine ExoViP can be the only solution to improve the compositional  visual reasoning methods

In this work, we identify two key types of errors in existing compositional methods: planning errors and module errors. To address these errors, we introduce an innovative verification framework \model. This framework verifies the correctness of vision module predictions. It corrects module errors by calibration and refines the planning process through tree searching. During this process, it considers both verification scores and the self-correctness of LLM. Applying the \model to two existing models, we achieve performance improvements across five different tasks. The results reinforce the promise and potential of \model on various open-world compositional reasoning tasks, marking an important milestone in the realm of multi-modal tasks involving complex reasoning.


\paragraph{Acknowledgments}
The authors thank the reviewers for their insightful suggestions to improve the manuscript. This work presented herein is supported by the National Natural Science Foundation of China (62376031).


\clearpage

% \section*{Impact Statement}
% ethical aspects and social consequences
% This paper contributes to advancing the field of compositional visual reasoning and illuminates potential avenues for future research. We have ensured that all baselines and datasets are utilized in accordance with their respective licenses. 
% While there are potential societal impacts stemming from our work, we do not believe any specific consequences necessitate particular emphasis in this context.

\bibliography{colm2024_conference}
\bibliographystyle{colm2024_conference}

% \appendix

\clearpage
{\bf\huge Appendices \bigskip}

\begin{appendices}
\label{sec:appendix}

\DoToC

% \section{Appendix}

\section{Exoskeleton Algorithm}
\label{supp:algo}

We demonstrate the overall algorithm of our method in Algorithm~\ref{alg:exovip}. There are mainly two parts: step-by-step verification and exploration with reasoning trace. To be more specific, we fuse the self-correctness ability of LLM into the procedure of tree-based reasoning trace searching, which has shown potential in calibrating the effectiveness of the searching algorithm.

\begin{algorithm}[h!]
  \SetAlgoLined
  \DontPrintSemicolon
  \SetKwComment{Comment}{/* }{ */}
\caption{Exoskeleton Algorithm}
\label{alg:exovip}
\KwIn{start step ($e_0$), goal node ($g$), scaling factor ($\tau$), verification size ($K$), rank size ($P$)}
\KwOut{Verified reasoning trace and intermedia results}
\SetKwFunction{ExoViP}{ExoViP}
% \ExoViP{$e_0$, $g$, $\beta$}{}
$openList$ $\gets$ $e_0$\;
$closedList$ $\gets$ $empty\ list$\;
$path \gets empty\ list$\;
\While{open list is not empty}{
    $sort(openList,\ key=e_s)$\;
    Select top $K$ steps from $openList$ and put it in $closedList$ and empty $openList$\;
    $rank(closedList, \ key=LLM(e))$\;
    Select top $P$ steps to update $closedList$\;
    \For{$e\ in\ closedList$}{
        \eIf{$e\ is\ g$}{
            $path.add(e)$\;
            return $path$\;
        }{
            $openList.add(e.next)$\;
        }}
    \For{$e\ in\ openList$}{ 
        $e_s = avg(e_s^{item} - e_n^{item}, e_s^{cap} - e_n^{cap}, e_s^{vqa}-e_n^{vqa})$\;
        $e \gets Verify(NORM(e_s, \tau), e)$
    }
}
\end{algorithm}

\section{Proof-of-concept Pilot Experiments}
\label{supp:proof}

\begin{figure}[h]
    \centering
    \includegraphics[width=.7\linewidth]{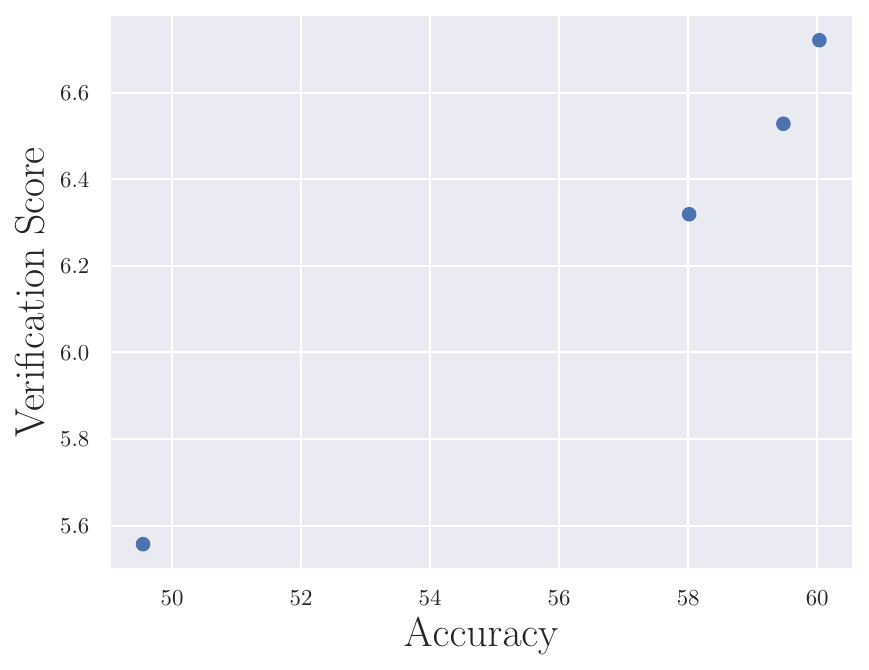}
    \caption{Accuracy on GQA positively correlates with the verification scores.}
    \label{fig:corel}
\end{figure}

To evaluate the effectiveness of the verification modules, we try to find the relationship between verification scores and accuracy. All experiments are applied to the GQA dataset. We first disturb the examples in the demonstrations to get different plan results and corresponding verification scores. Specifically, we change the order of examples and select different portions of examples with four settings. After evaluation, we calculate the mean of verification scores of all steps. As is shown in~\cref{fig:corel}, we are delighted to find the verification scores positively contribute to final accuracy. However, the trend is decreasing, which means when the verification scores increase to a certain extent, higher verification scores do little contribution to the final accuracy.

\section{Enhanced Verification through TRS}

\begin{figure}[H]
    \centering
	\includegraphics[width=.5\linewidth]{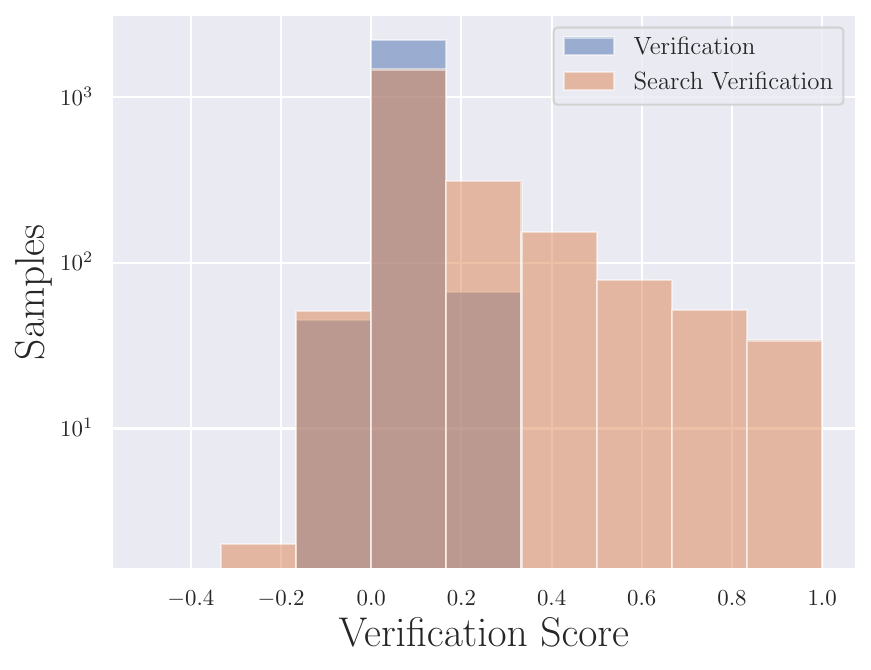}
    \vspace{-0.9em}
    \captionof{figure}{Distribution of verification scores w. and w/o trace searching. }
    \vspace{-0.5em}
    \label{fig:dist} 
\end{figure}

In ~\ref{fig:dist}, we present a graphical representation illustrating the relationship between the trace-searching strategy and the verification score. The x-axis quantifies the verification score associated with each trace, while the y-axis denotes the number of traces corresponding to each score.

\begin{figure}[H]
    \centering
    \includegraphics[width=\linewidth]{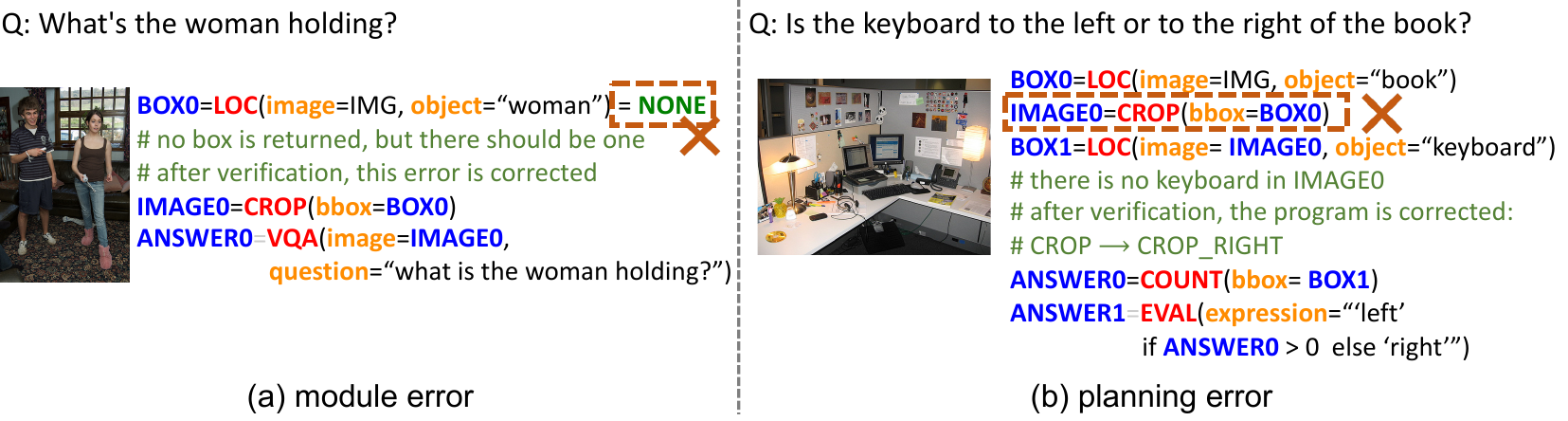}
    \caption{Existing methods suffer from two types of errors: (a) vision module prediction error and (b) LLM planning error. Our verification modules help correct the errors.}
    \label{fig:failure}
\end{figure}

\begin{figure}[H]
    \centering
    \includegraphics[width=\linewidth]{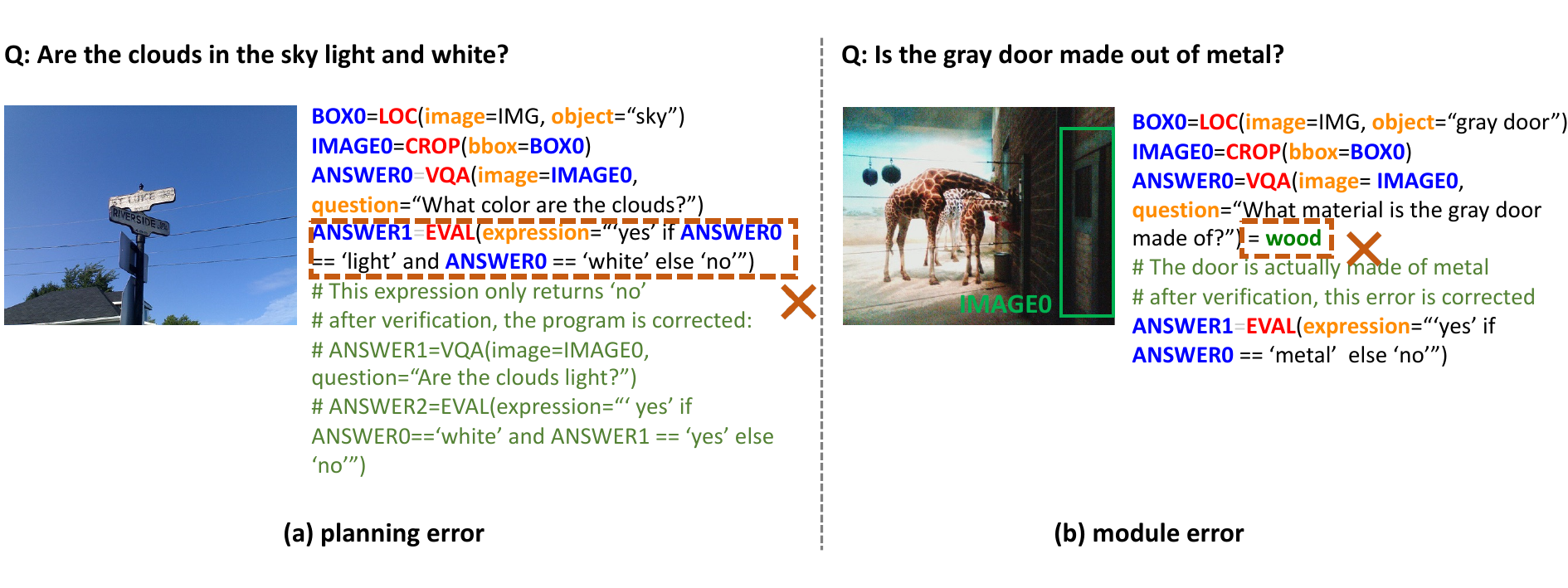}
    \caption{More examples of the two types of errors: (a) vision module prediction error and (b) LLM planning error.}
    \label{fig:failures}
\end{figure}

In~\cref{fig:failure,fig:failures}, we show examples of failure cases of the original \visprog. 

\begin{figure}[H]
    \centering
    \includegraphics[width=\linewidth]{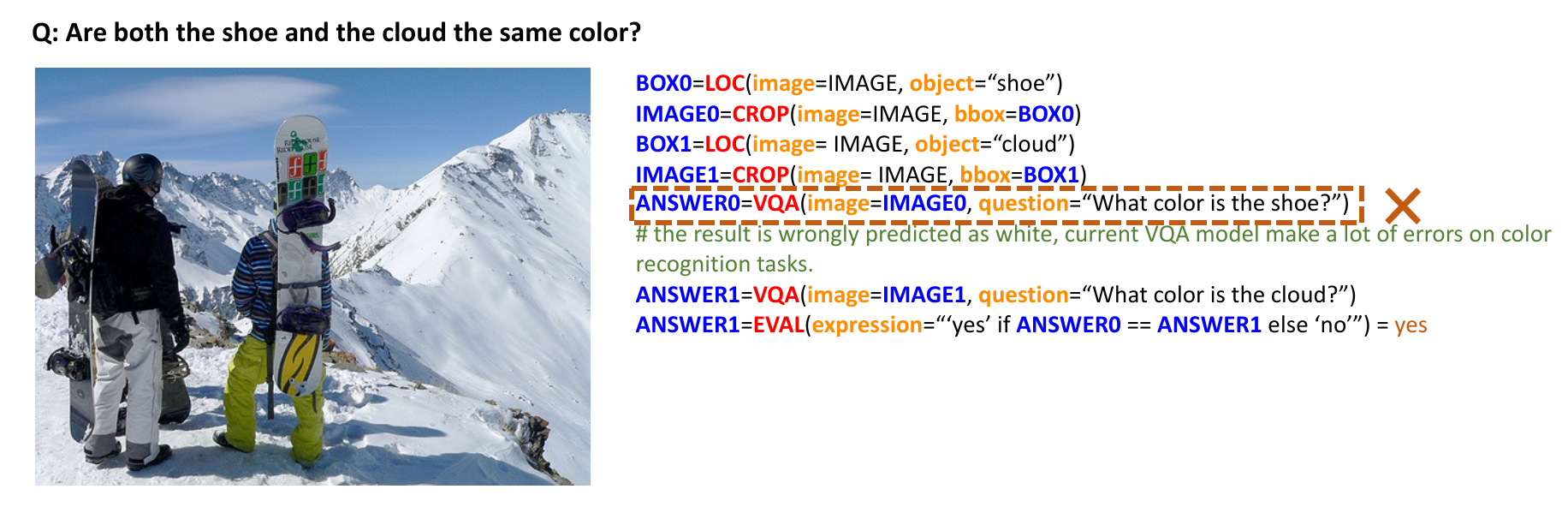}
    \caption{Common failure cases: some modules perform badly on certain tasks, \eg the VQA module performs poorly on color recognition tasks.}
    \label{fig:failurecase-module}
\end{figure}

\begin{figure}[H]
    \centering
    \includegraphics[width=\linewidth]{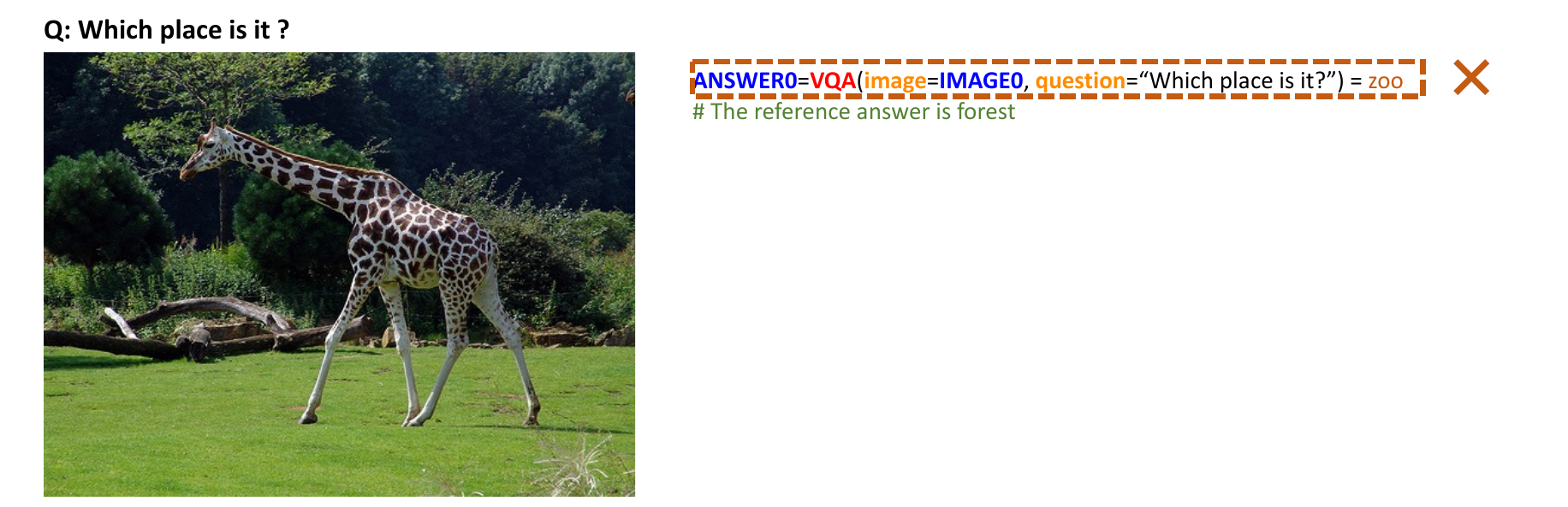}
    \caption{Common failure cases: some queries can not be decomposed into sub-tasks. Our method helps little with these non-decomposable queries.}
    \label{fig:failurecase-plan}
\end{figure}

% \section{\added{Results on open LLMs}}
% \label{supp:open-llm}
% \added{In this section, we present the results of applying our method to the open LLM. Specifically, we substituted GPT-3.5-turbo with LLama2-chat-13b. The outcome of this substitution is displayed in }\cref{tab:openllm}. We are thrilled to discover that our method can yield significant improvements in open LLM.

% \begin{table}[H]
%     \centering
%     \captionof{table}{\added{Results of open LLMs on GQA}}
%     \label{tab:openllm}
%     \begin{tabular}{lc}
%         \toprule
%          \multicolumn{1}{l}{\bf Methods}  &\multicolumn{1}{c}{\bf Accuracy} \\
%         \midrule
%         \visprog (GPT-3.5-turbo)             & 57.41 \\ 
%         \model (GPT-3.5-turbo)            & 61.49 \\
%         \visprog (Llama-2-13b-chat)        & 46.41 \\ 
%         \model ((Llama-2-13b-chat)) & 54.45 \\ \bottomrule
%     \end{tabular}
% \end{table}

\section{Efficiency Analysis}
\label{supp:efficiency}
We present the average inference time on the GQA dataset. Generally, the temporal expenditure of our tree-based search method significantly surpasses that of VisProg. However, the majority of the time is consumed by the call of the OPENAI API, an issue we posit is intrinsic to analogous works \cite{DBLP:journals/corr/abs-2305-10601,DBLP:journals/corr/abs-2309-17179,DBLP:journals/corr/abs-2310-04406}. When compared to Depth First Search/ Breadth First Search \cite{DBLP:journals/corr/abs-2305-10601} or Monte Carlo Tree Search \cite{DBLP:journals/corr/abs-2309-17179,DBLP:journals/corr/abs-2310-04406}, we assert that our beam search-based method can achieve an optimal equilibrium between efficiency and effectiveness. In addition, when comparing with End-to-End model, we find that the most significant contributor to the overall time cost is the Planning Time. We have determined that this delay is largely attributable to Internet latency, as our system utilizes the GPT-3.5-turbo API. We are confident that this latency can be mitigated by deploying the LLM locally, which would reduce the dependency on network response times. Additionally, we are exploring ways to enhance the parallelism of our system's submodules, which we believe will further improve efficiency.

% \begin{wraptable}{r}{\textwidth}
%     \centering
%     \captionof{table}{Average Inference Time on the GQA Dataset}
%     \label{tab:efficiency}
%     \begin{tabular}{lccc}
%         \toprule
%          \multicolumn{1}{l}{\bf Methods}  &\multicolumn{1}{c}{\bf Total Inference Time (s)} &\multicolumn{1}{c}{\bf Planning Time (s)} &\multicolumn{1}{c}{\bf Module Inference Time (s)} \\
%         \midrule
%         BLIP2-Flant5-xxl             & 0.17 & - & 0.17 \\
%         LLaVA-1.5-7B             & 0.45 & - & 0.45 \\
%         \visprog              & 1.59 & 1.10 & 0.49 \\ 
%         \model             & 4.32 & 3.64 & 0.68 \\ \bottomrule
%     \end{tabular}
% \end{wraptable}

\begin{table}[H]
    \centering
    \captionof{table}{Average Inference Time on the GQA Dataset}
    \label{tab:efficiency}
    \begin{tabular*}{\linewidth}{lccc}
        \toprule
         \multicolumn{1}{l}{\bf Methods}  &\multicolumn{1}{c}{\bf Total Infer. Time (s)} &\multicolumn{1}{c}{\bf Planning Time (s)} &\multicolumn{1}{c}{\bf Module Infer. Time (s)} \\
        \midrule
        BLIP2-Flant5-xxl             & 0.17 & - & 0.17 \\
        LLaVA-1.5-7B             & 0.45 & - & 0.45 \\
        \visprog              & 1.59 & 1.10 & 0.49 \\ 
        \model             & 4.32 & 3.64 & 0.68 \\ \bottomrule
    \end{tabular*}
\end{table}

% \section{More experiments on grounding}
% \label{supp:exp-refcocog}

\section{Implementation details}
\label{supp:implementation}

\subsection{Visual modules.}
\label{supp:visual-module}

\begin{figure}[H]
    \centering
    \includegraphics[width=\linewidth]{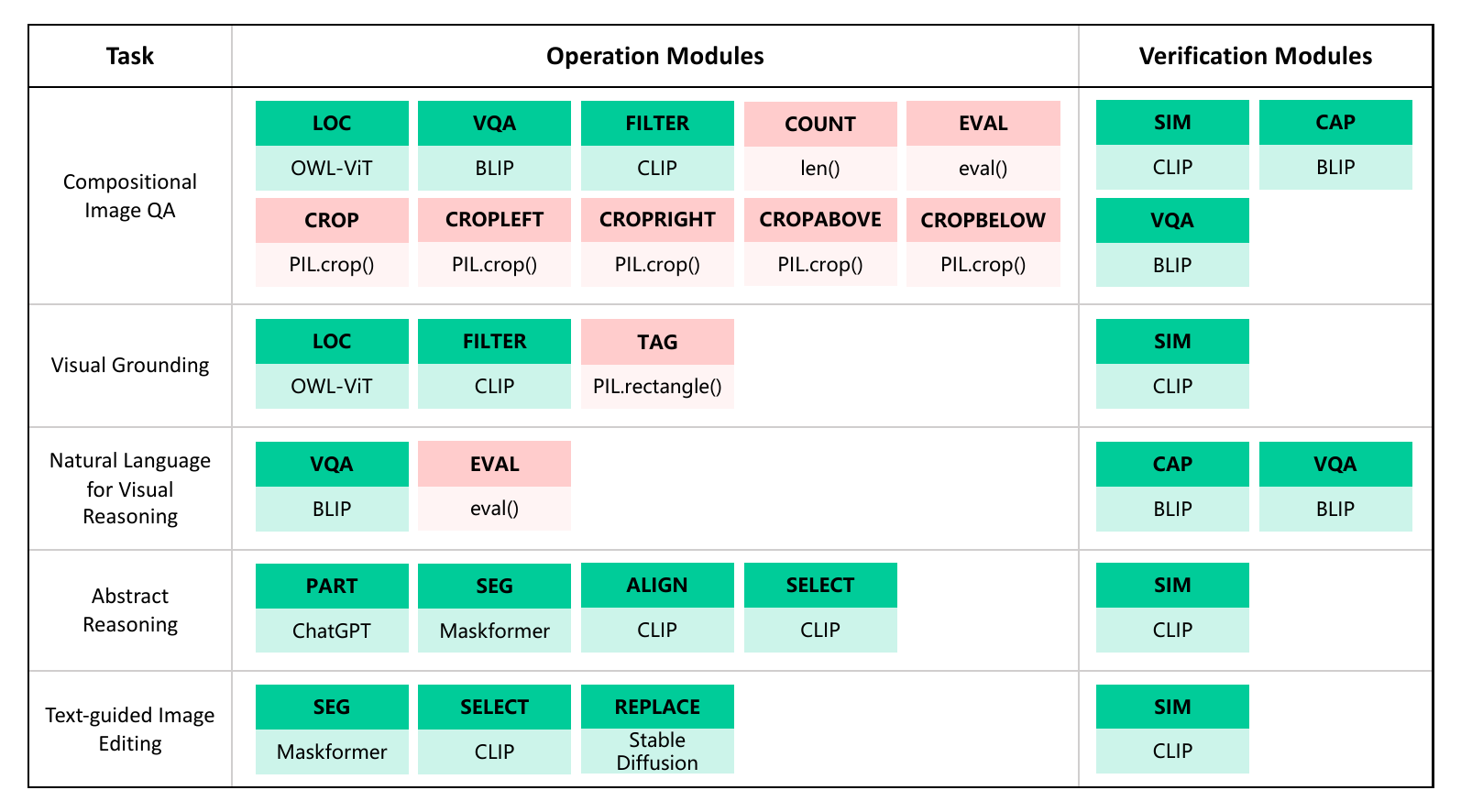}
    \caption{The neural modules (green) and symbolic modules (pink) used in our experiments.}
    \label{fig:modules}
\end{figure}

We summarize the operation modules and the verification modules of different tasks in ~\cref{fig:modules}. In practice, the candidate neural modules include OWL-ViT~\citep{DBLP:journals/corr/abs-2205-06230}, CLIP~\citep{Radford2021LearningTV}, BLIP~\citep{Li2022BLIPBL}, ChatGPT, MaskFormer~\citep{DBLP:conf/nips/ChengSK21}, Stable Diffusion~\citep{DBLP:conf/cvpr/RombachBLEO22}. In order to validate the effectiveness of our method and eliminate the benefits of external knowledge such as more advanced vision-language models which are trained on larger datasets. Both operation modules and verification modules are selected from the same candidate neural module sets. In other words, not all modules are verified on the mixture of all three types of modules.

\subsection{LLM Prompts}
\label{supp:prompts}
% \zhuowan{example of the prompts can be put here. Especially for the LLM correctness }
We demonstrate the prompt for self-correctness of all five tasks.
\begin{figure}[H]
    \centering
    \includegraphics[width=\linewidth]{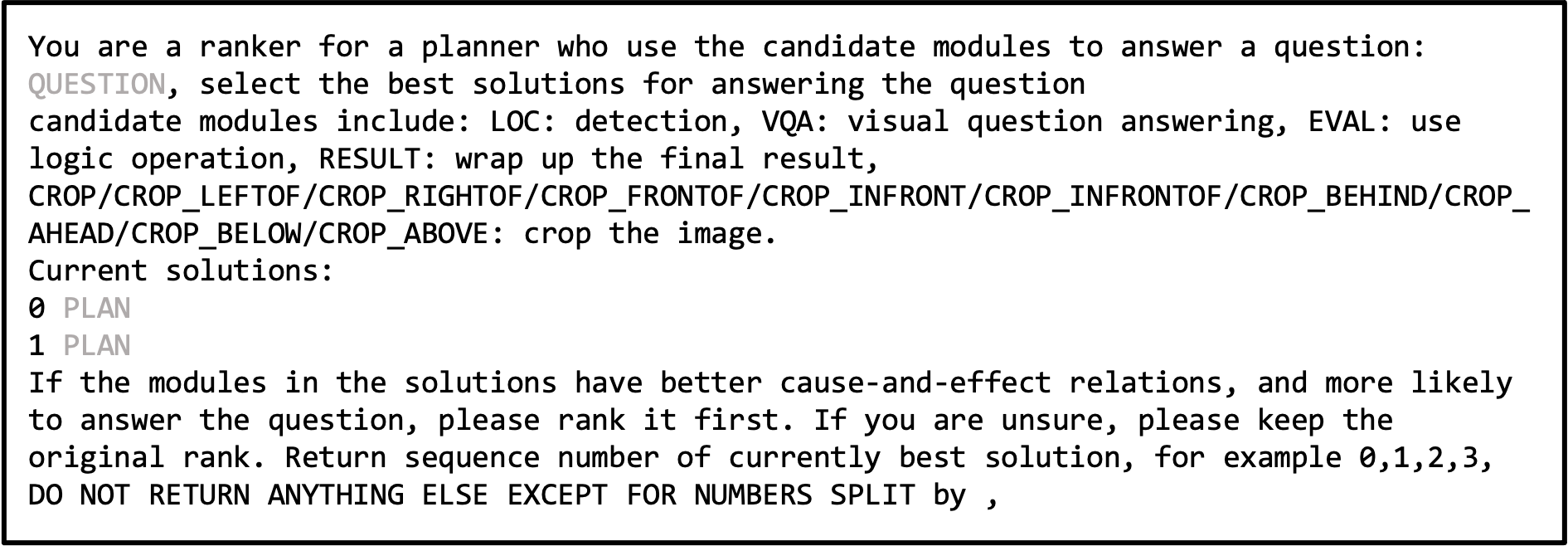}
    \caption{Self-correctness prompt of compositional question answering.}
    \label{fig:prompt-gqa}
\end{figure}
\begin{figure}[H]
    \centering
    \includegraphics[width=\linewidth]{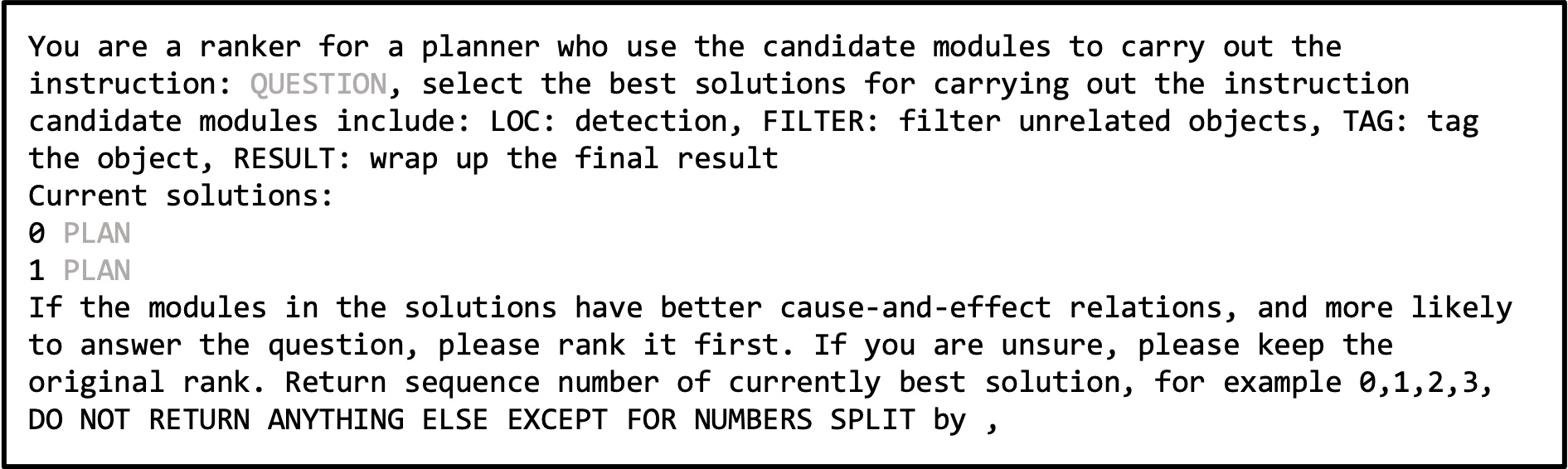}
    \caption{Self-correctness prompt of visual grounding.}
    \label{fig:prompt-refcoco}
\end{figure}
\begin{figure}[H]
    \centering
    \includegraphics[width=\linewidth]{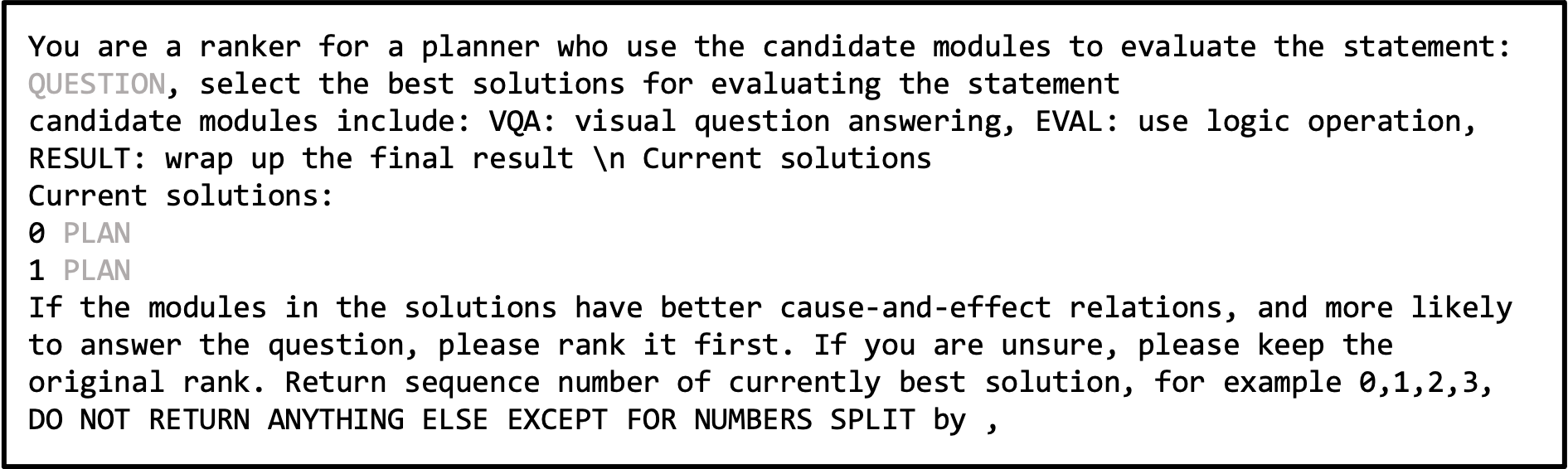}
    \caption{Self-correctness prompt of natural language for visual reasoning.}
    \label{fig:prompt-nlvr}
\end{figure}
\begin{figure}[H]
    \centering
    \includegraphics[width=\linewidth]{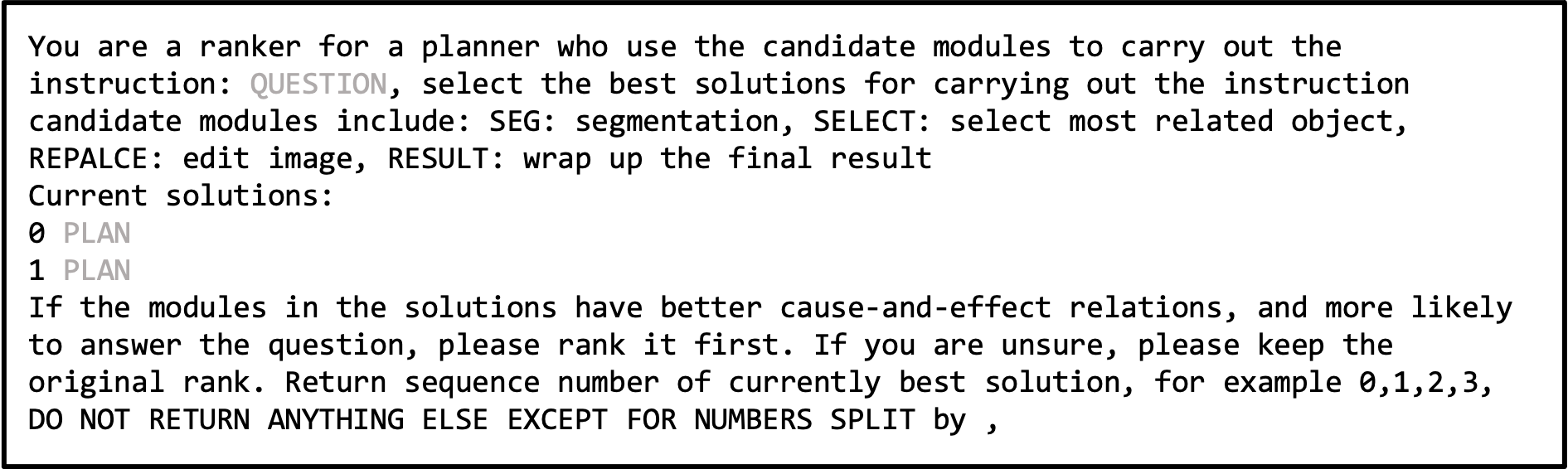}
    \caption{Self-correctness prompt of text-guided image editing.}
    \label{fig:prompt-magicbrush}
\end{figure}
\begin{figure}[H]
    \centering
    \includegraphics[width=\linewidth]{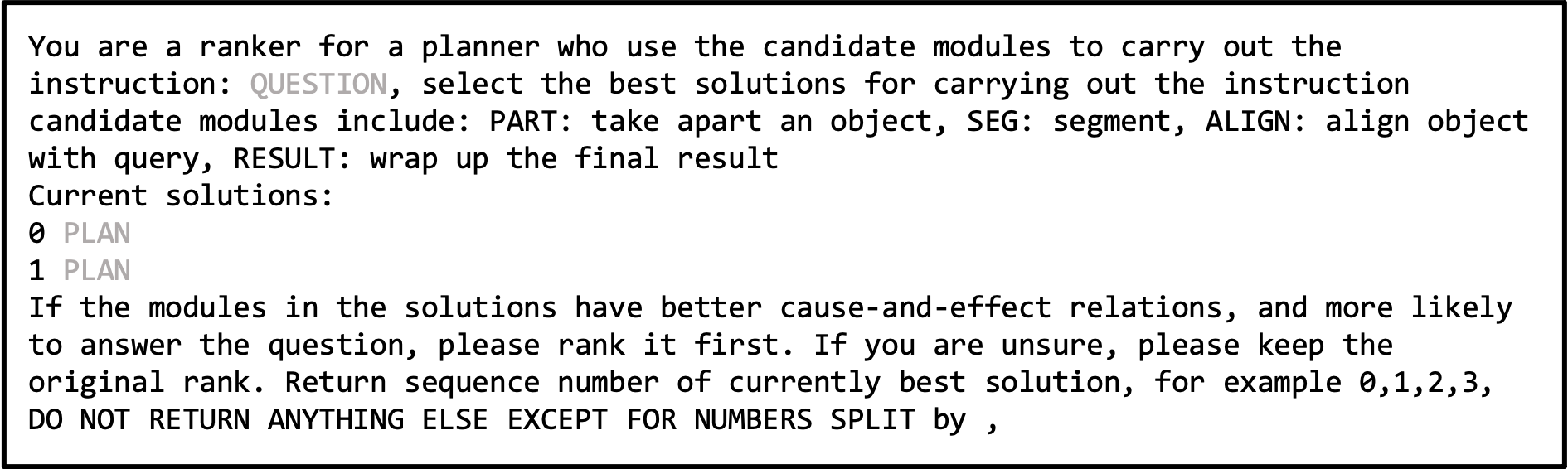}
    \caption{Self-correctness prompt of visual abstract reasoning.}
    \label{fig:prompt-kilogram}
\end{figure}

\subsection{Details of Visual Abstract Reasoning} 
\label{supp:VAR_detail}
\begin{figure}[H]
    \centering
    \includegraphics[width=\linewidth]{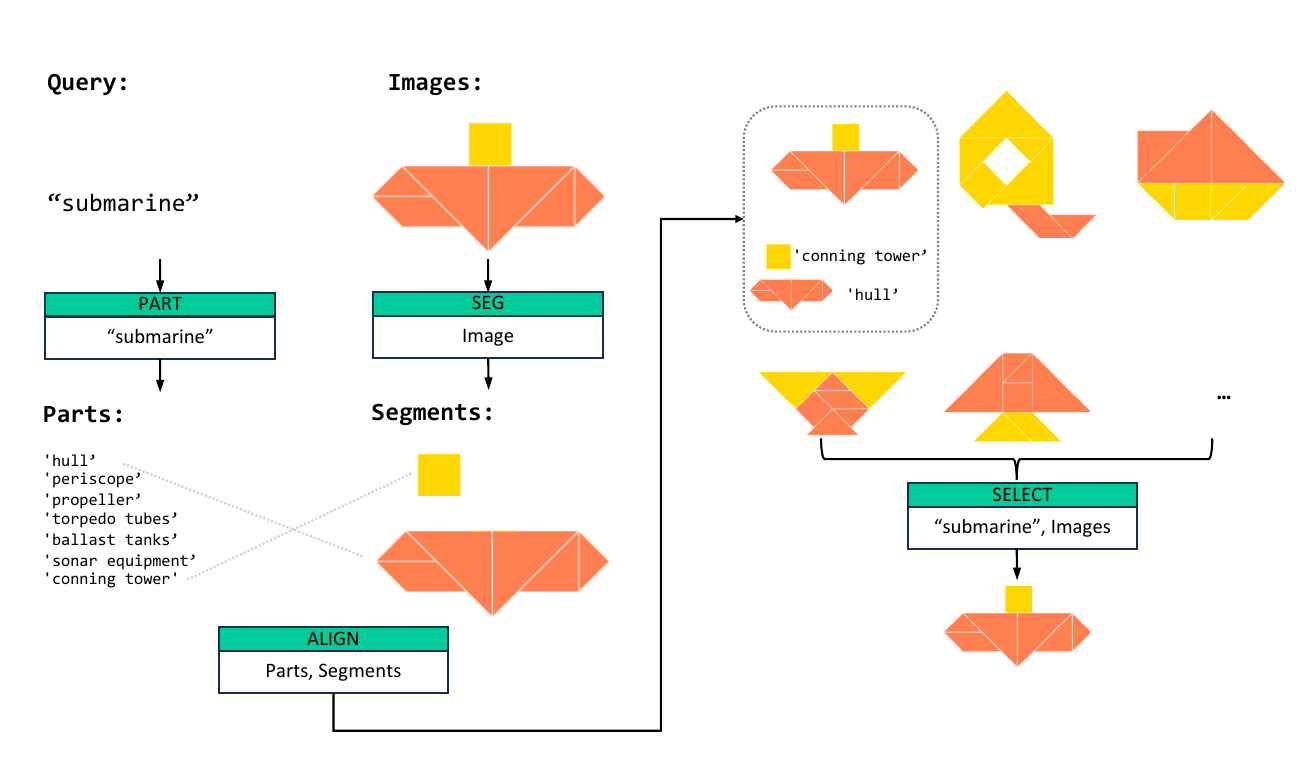}
    \caption{Implementation of abstract reasoning.}
    \label{fig:ar}
\end{figure}
% \begin{wrapfigure}{r}{.45\textwidth}
%     \centering
%     \includegraphics[width=\linewidth]{figures/cases/exovip_qualitative_kilogram.pdf}
%     \caption{Implementation of \model on abstract reasoning.}
%     \label{fig:kilogram}
% \end{wrapfigure}

In ~\cref{fig:ar}, we demonstrate our implementation of compositional methods on KILOGRAM dataset. Given an image, we segment it into several parts. At the same time, we adopt LLM to parse the query to several components. After that, we match the visual and textual components by their semantic similarity. Finally, we take the alignment score to retrieve the best matched image.

\begin{figure}[H]
    \centering
    \includegraphics[width=.8\linewidth]{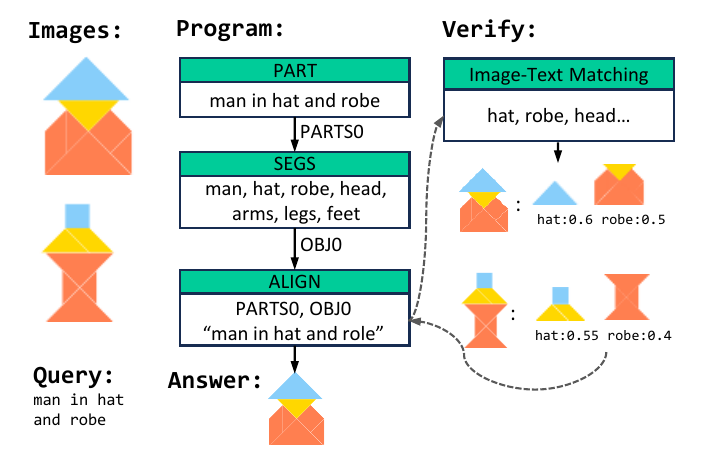}
    \caption{Implementation of \model on abstract reasoning.}
    \label{fig:kilogram}
\end{figure}

\subsection{Implementation details}
In practice, for the verification modules, we set the $\tau$ as $2.0$ for $LOC$ module,  $1.5$ for $SELECT$ module, and $ALIGN$ module,  $1.2$ for other modules. For the negative sampling strategy, we select words sharing semantic similarity less than $0.5$ to construct the semantic opposite vocabulary and randomly sample one semantic opposite for each answer.   In the searching process, we set up  $K$ as $4$, and $P$ as $2$. To improve the efficiency of our search algorithm, we set the branching factor as $3$. To make the comparison fair, we use the same or fewer examples in the prompts for our methods, and select the verification modules from the operation modules. We apply our experiments on NVIDIA A100 GPU and NVIDIA 3090Ti GPU. 

% \zhuowan{Put in details, like values of $\tau, K, P$. Hyperparameters for decoding etc.}

\section{Qualitative study.}
% \zhuowan{Failure cases and qualitative examples of our method.}

\subsection{Qualitative examples}

We additionally exhibit more examples that can be improved by our method. As is shown in these examples, all five types of tasks could be further improved by our framework.

\begin{figure}[H]
    \centering
    \includegraphics[width=\linewidth]{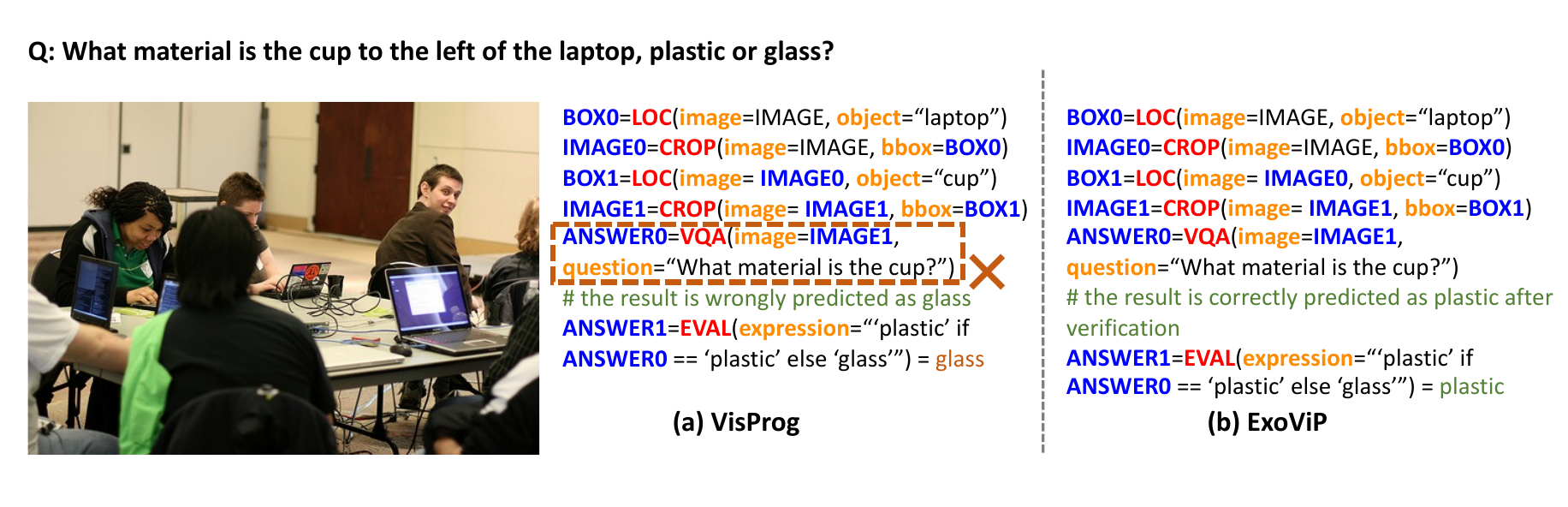}
    \caption{Qualitative study for GQA.}
    \label{fig:case-gqa}
\end{figure}

\begin{figure}[H]
    \centering
    \includegraphics[width=\linewidth]{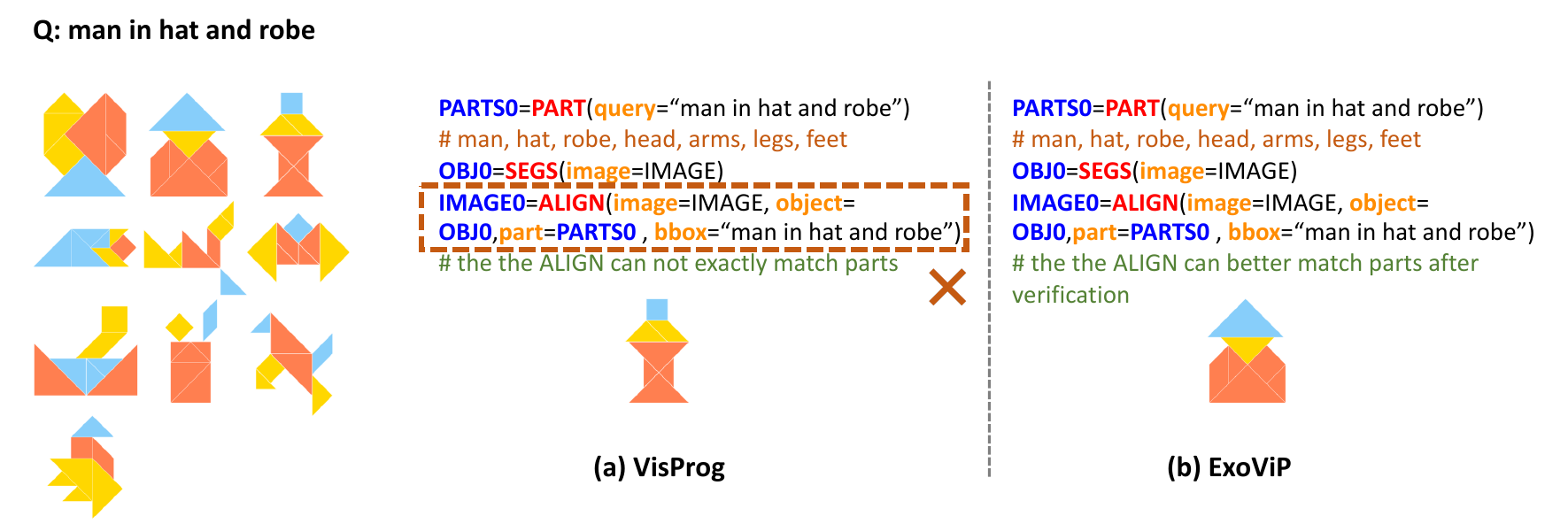}
    \caption{Qualitative study for KILOGRAM.}
    \label{fig:case-kilogram}
\end{figure}

\begin{figure}[H]
    \centering
    \includegraphics[width=\linewidth]{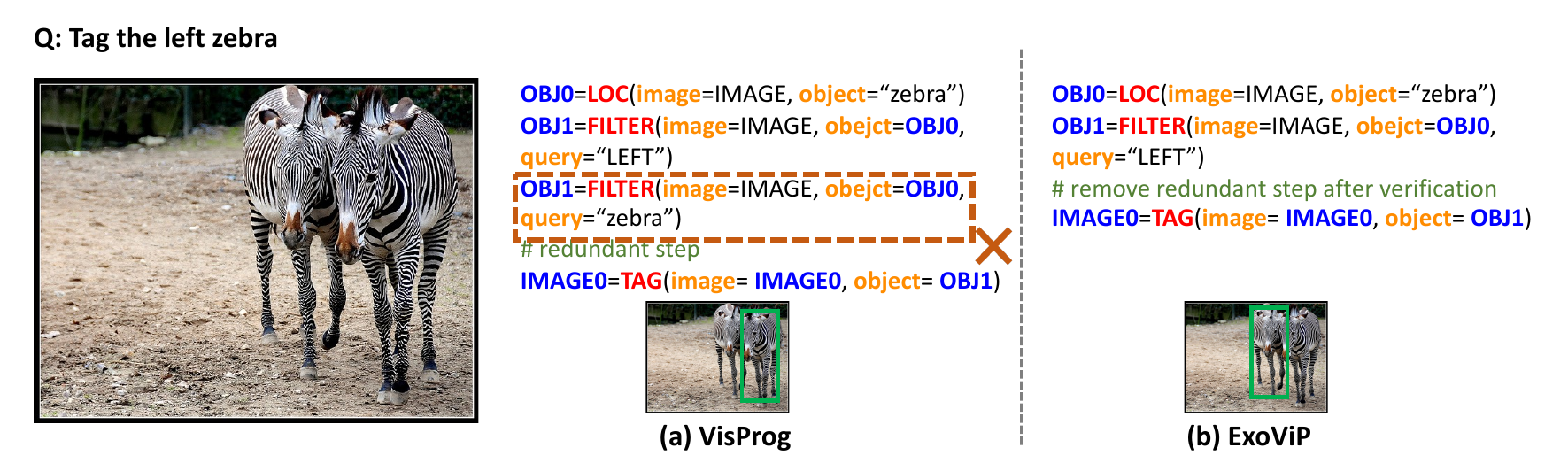}
    \caption{Qualitative study for RefCOCO.}
    \label{fig:case-refcoco}
\end{figure}

\begin{figure}[H]
    \centering
    \includegraphics[width=\linewidth]{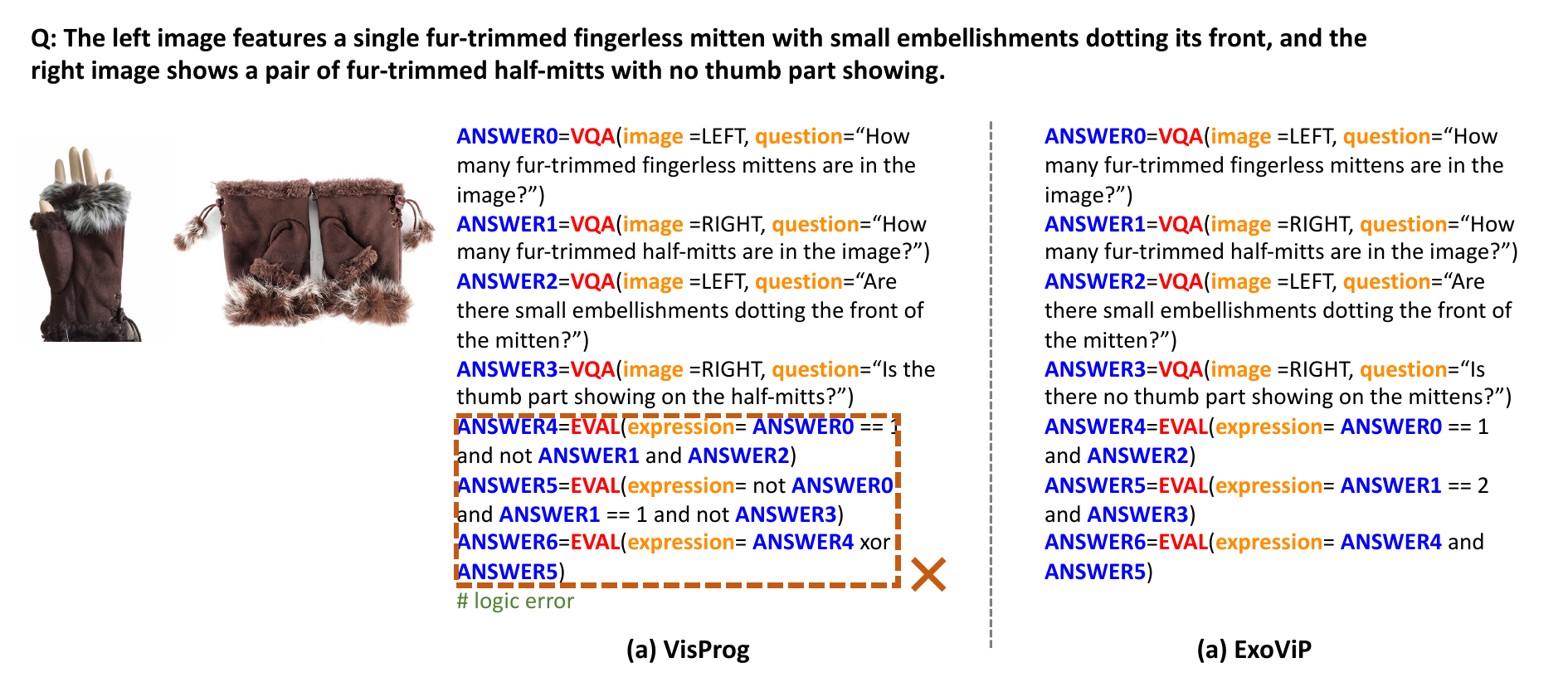}
    \caption{Qualitative study for NLVR2.}
    \label{fig:case-nlvr}
\end{figure}

% \subsection{Failure cases}
% \label{supp:failure-cases}

% In this part, we showcase common failure cases, which still exist in our method. We find the majority of these instances originate from the VQA module.

% \begin{figure}[H]
%     \centering
%     \includegraphics[width=\linewidth]{figures/cases/exovip_case_failure_module.pdf}
%     \caption{Common failure cases: some modules perform badly on certain tasks, \eg the VQA module performs poorly on color recognition tasks.}
%     \label{fig:failurecase-module}
% \end{figure}

% \begin{figure}[H]
%     \centering
%     \includegraphics[width=\linewidth]{figures/cases/exovip_case_failure_plan.pdf}
%     \caption{Common failure cases: some queries can not be decomposed into sub-tasks. Our method helps little with these non-decomposable queries.}
%     \label{fig:failurecase-plan}
% \end{figure}

\end{appendices}

\end{document}

%% file: config.tex
\usepackage[pagebackref,breaklinks,colorlinks=true, citecolor=blue,linkcolor=black]{hyperref}
\usepackage{url}

\usepackage{enumitem}
\usepackage{graphicx}
\graphicspath{figures/}
\usepackage{multirow}
\usepackage{extarrows}
\usepackage{acronym}
\usepackage{xspace}
\usepackage[capitalize]{cleveref}
\usepackage[tableposition=top]{caption}
\usepackage{wrapfig}
\usepackage{booktabs}
\usepackage{titletoc}
\usepackage{appendix}
\usepackage{pifont}

\usepackage[ruled]{algorithm2e}
\usepackage{bbm}
\usepackage{float}
\usepackage{xcolor,colortbl}
\usepackage{makecell}

\usepackage{diagbox}
\usepackage{verbatim}
\usepackage{framed}
\usepackage[most]{tcolorbox}
\usepackage{amsmath}
\usepackage[misc]{ifsym}
\usepackage{fontawesome5}
\crefname{section}{Sec.}{Secs.}
\Crefname{section}{Section}{Sections}
\Crefname{table}{Table}{Tables}
\crefname{table}{Tab.}{Tabs.}
\Crefname{figure}{Figure}{Figures}
\crefname{figure}{Fig.}{Figs.}

\newcommand\DoToC{%
  \startcontents
  \printcontents{}{1}{\textbf{\Large Table of Contents}\vskip5pt\hrule\vskip3pt}
  \vskip3pt\hrule\vskip5pt
}

\makeatletter
\renewcommand{\paragraph}{%
  \@startsection{paragraph}{4}%
  {\z@}{0ex \@plus 0ex \@minus 0ex}{-1em}%
  {\hskip\parindent\normalfont\normalsize\bfseries}%
}
\makeatother

\makeatletter
\DeclareRobustCommand\onedot{\futurelet\@let@token\@onedot}
\def\@onedot{\ifx\@let@token.\else.\null\fi\xspace}
\def\eg{\emph{e.g}\onedot} 
\def\ie{\emph{i.e}\onedot}

\def\wrt{w.r.t\onedot} 

\makeatother

\acrodef{llm}[LLM]{Large Language Model}
\acrodef{kg}[KG]{knowledge graph}

%% file: math_commands.tex
\usepackage{amsmath,amsfonts,bm}

\def\eqref#1{equation~\ref{#1}}
\def\1{\bm{1}}

\DeclareMathAlphabet{\mathsfit}{\encodingdefault}{\sfdefault}{m}{sl}
\SetMathAlphabet{\mathsfit}{bold}{\encodingdefault}{\sfdefault}{bx}{n}